\documentclass[11pt]{article}
\usepackage[margin=1in]{geometry}
\usepackage{amsmath,amssymb}
\usepackage{booktabs}
\usepackage{longtable}
\usepackage{float}
\usepackage{graphicx}
\usepackage{xcolor}
\usepackage{fontawesome5} 
\usepackage{natbib}
\usepackage{tikz}
\usetikzlibrary{positioning,arrows.meta,decorations.pathreplacing,calc}
\usepackage{lmodern} 
\usepackage{fancyhdr} 
\usepackage[htt]{hyphenat} 
\definecolor{tabBlueF}{HTML}{B5D4F4}  \definecolor{tabBlueS}{HTML}{185FA5}
\definecolor{tabTealF}{HTML}{9FE1CB}  \definecolor{tabTealS}{HTML}{0F6E56}
\definecolor{tabGrayF}{HTML}{D3D1C7}  \definecolor{tabGrayS}{HTML}{5F5E5A}
\definecolor{tabPurpF}{HTML}{AFA9EC}  \definecolor{tabPurpS}{HTML}{534AB7}
\definecolor{tabAmberF}{HTML}{FAC775} \definecolor{tabAmberS}{HTML}{854F0B}
\definecolor{tabGreenF}{HTML}{C0DD97} \definecolor{tabGreenS}{HTML}{3B6D11}
\definecolor{tabPinkF}{HTML}{F4C0D1}  \definecolor{tabPinkS}{HTML}{993556}
\definecolor{panF}{HTML}{F1EFE8}      \definecolor{panS}{HTML}{D3D1C7}
\definecolor{badgeBlue}{HTML}{185FA5} \definecolor{tealMid}{HTML}{5DCAA5}
\definecolor{tealBg}{HTML}{E1F5EE}    \definecolor{purpBg}{HTML}{EEEDFE}
\definecolor{amberBg}{HTML}{FAEEDA}    \definecolor{greenBg}{HTML}{EAF3DE}
\definecolor{fadedS}{HTML}{B4B2A9}
\definecolor{stepBg}{HTML}{F7FAFC}
\definecolor{hfyellow}{HTML}{FFD21E} 
\usepackage[colorlinks=true,linkcolor=blue,citecolor=blue,urlcolor=blue]{hyperref}

\newcommand{\model}{TabAbax}

\begin{document}

\fancypagestyle{firstpage}{%
  \fancyhf{}%
  \renewcommand{\headrulewidth}{0pt}%
  \renewcommand{\footrulewidth}{0.4pt}%
  \fancyfoot[C]{\small
  $^{*}$ Corresponding authors: \texttt{jingjing.suo@brgroup.com}, \texttt{zhengli.wang@brgroup.com}}%
}
\thispagestyle{firstpage}
\begin{center}
{\LARGE \model{}: A Tabular Foundation Model Technical Report}\\[6pt]
{\large Zhixiong Cheng, Xiaoyun Gao, Shiyuan Han, Jingjing Suo\textsuperscript{*}, Zhengli Wang\textsuperscript{*}}\\[2pt]
{\normalsize Bairong Inc.AI, Beijing, China}\\[5pt]
{\large 2026-10-31}\\[6pt]
{\small
\faGithub\ \ \url{https://github.com/tababax-research/TabAbax}\\[3pt]
{\textcolor{hfyellow}{\faSmile}}\ \ \url{https://huggingface.co/tababax-ai/TabAbax}}
\end{center}

\vspace{2pt}
\begin{center}\textbf{Abstract}\end{center}
\noindent
Tabular foundation models (TFMs) enable in-context prediction on
unseen tabular datasets without task-specific training. Mechanistic
work has shown that increasing the depth of the ICL Transformer
yields  weak gains,
leaving the main headroom for gains in the synthetic pretraining
prior and column--row encoder upstream of the ICL backbone. We
introduce \model{}, a TFM built on TabICLv2 that focuses all
modifications on these two upstream components. On the
prior side, we further improve the quality and complexity of the
synthetic pretraining prior. On the encoder side, a
two-pass column--row stack strengthens cross-feature interactions
before row compression, using learned Fourier scalar encoding from
TabFM and Causilo. Evaluations on BeyondArena and TabArena show
strong regression-centric gains: \model{} reaches 1{,}238 Elo on
the 142-dataset BeyondArena suite (10th of 41 evaluated methods),
1{,}676 Elo on TabArena's overall 51-dataset suite (11th of 99
methods, $+$115 over TabICLv2), and 2{,}024 Elo on TabArena's
regression benchmark (7th of 99
methods, $+$336 over TabICLv2, $+$21 over TabFM). A prior-only
ablation confirms the
attribution: the
improved prior alone yields $+$228 Elo for regression,
demonstrating that prior quality is the
dominant knob for improving regression performance.

\begin{figure}[H]
\centering
\includegraphics[width=0.85\textwidth]{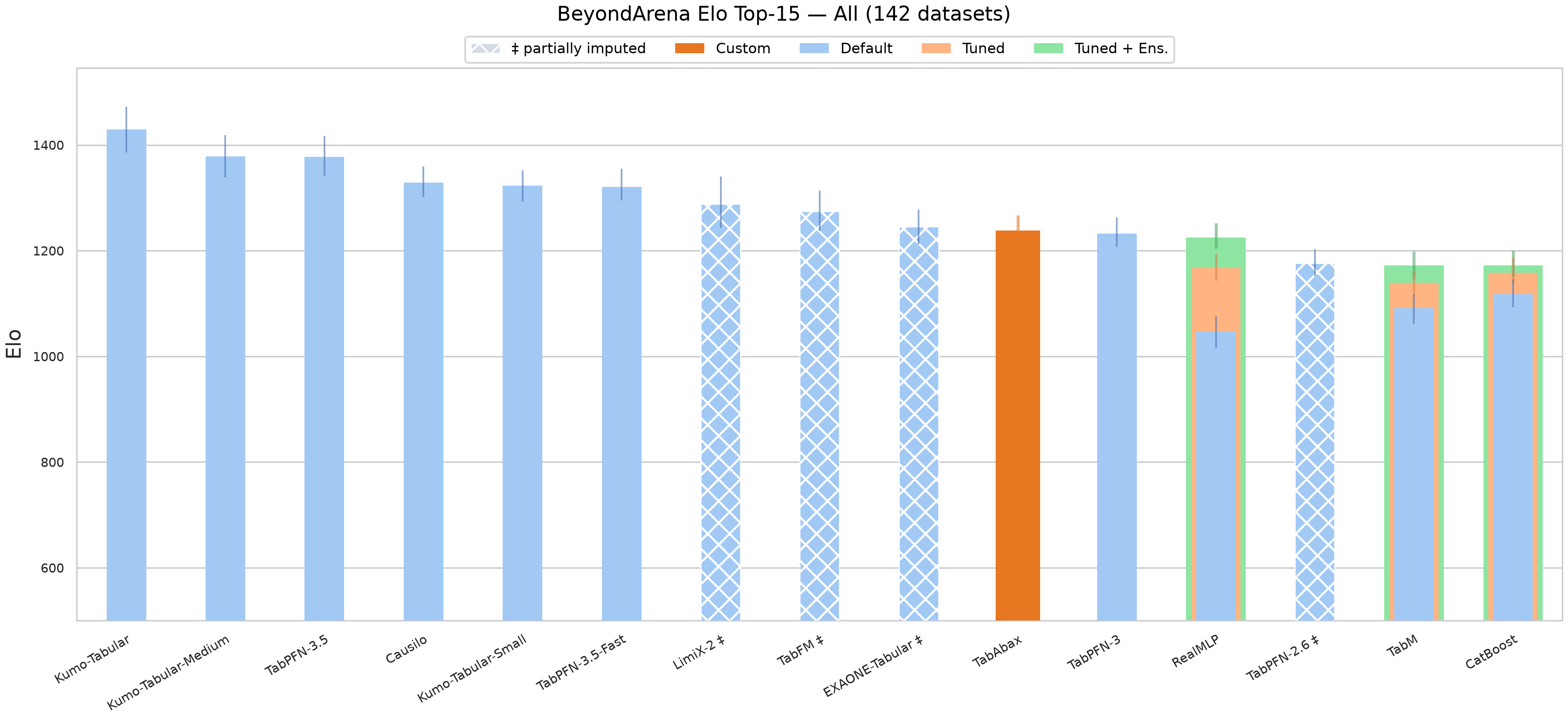}
\caption{BeyondArena overall Elo across the full 142-dataset suite
(98 classification, 44 regression; top-15 leaderboard).
\model{} (dark orange bar) reaches an Elo rating of 1{,}238, ranking 10th of
the 41 evaluated methods, $+$68 over the TabICLv2 baseline (1{,}170).
Methods marked $\ddagger$ are partially imputed.}
\label{fig:beyondarena-all-top15}
\end{figure}

\begin{figure}[H]
\centering
\includegraphics[width=0.85\textwidth]{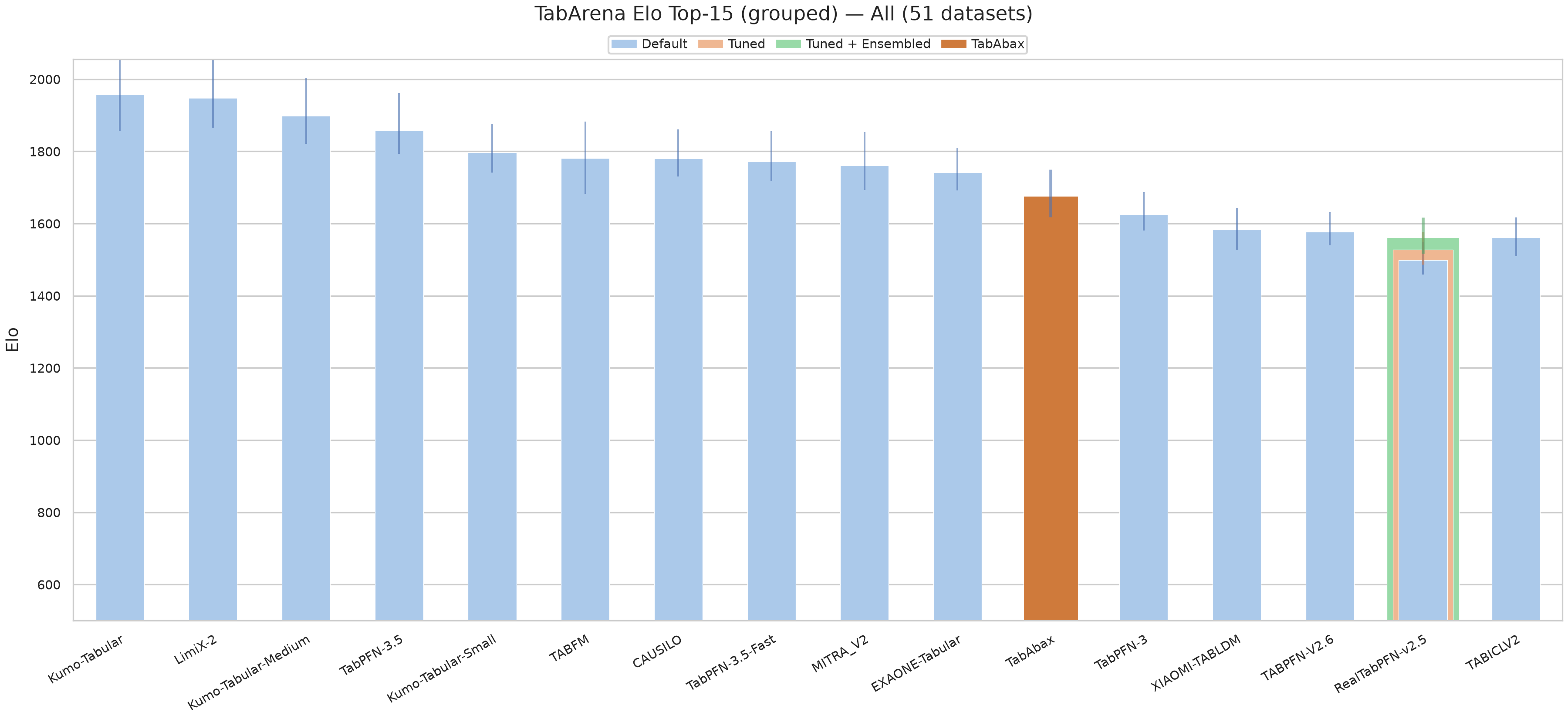}
\caption{TabArena overall Elo across all 51 datasets (top-15 grouped
leaderboard). \model{} (dark orange bar) reaches
an Elo rating of 1{,}676, ranking 11th of all
99 evaluated methods, $+$115
over the TabICLv2 official default baseline (1{,}561) and $+$50 over
TabPFN-3 (1{,}626). The full 99-method table is in
Table~\ref{tab:app-overall}.}
\label{fig:tabarena-overall}
\end{figure}

\section{Introduction}
Structured data is a foundational modality for scientific analysis and
operational decision-making across a wide range of domains, including
healthcare, finance, logistics, and manufacturing. Unlike language or
perceptual data, tabular data organizes heterogeneous variables under
explicit schemas, preserving numerical scales, missingness patterns, and
categorical structure that are essential for reliable prediction. For
decades, prediction on tabular data has been dominated by dataset-specific
pipelines, particularly gradient-boosted trees
\citep{chen2016xgboost,ke2017lightgbm,prokhorenkova2018catboost,grinsztajn2022tree},
which
remain highly competitive across diverse tabular problems; deep
alternatives have only recently reached parity \citep{ye2025tabr}.

Tabular foundation models provide a different route: a single pretrained
model learns across a large distribution of synthetic or real tabular tasks
and directly performs prediction on unseen datasets through in-context
learning, using labeled examples from the target dataset as context rather
than retraining model parameters. Since TabPFN showed that a transformer
\citep{vaswani2017attention} pretrained on synthetic tabular tasks
performs strong prediction in a single forward pass
\citep{hollmann2023tabpfnpfn,muller2022pfns,garg2022transformers},
subsequent work has scaled the paradigm through richer priors and more
expressive encoders
\citep{hollmann2025tabpfnnature,grinsztajn2025tabpfn25,qu2025tabicl,qu2026tabiclv2,kong2026tabfm,nums2026causilo},
real-data pretraining
\citep{ma2024tabdpt,hosseinzadeh2026tabdptturbo,realtabpfn25},
and language-model-based routes that serialize tables as text or scale
transfer learning to open-table corpora
\citep{hegselmann2023tabllm,gardner2024tabpfnlt}; Section~\ref{sec:related}
reviews these lines in detail.

Recent progress in tabular foundation models has come increasingly
from jointly scaling training data, model capacity, and inference
strategies. We take a complementary direction: rather than scaling,
we improve what feeds the ICL stack. A mechanistic study of tabular
foundation models
\citep{bilos2026mechanistic} supports this choice: architecturally
distinct models converge in accuracy and implement simple
similarity-based readouts, and many ICL layers carry little
load---the remaining headroom lies upstream, in the synthetic prior
and the column--row encoder, rather than in the ICL transformer
itself.
The synthetic SCM priors used for pretraining leave clear room for
improvement, and our prior-side changes concentrate on data quality
and complexity, acting across the entire generation pipeline: we enforce
strict one-hot width matching so that the emitted categories agree
exactly with the declared cardinalities; we generate class frequencies
through multiple distributions, e.g., a Dirichlet prior
$\boldsymbol{\pi} \sim \mathrm{Dirichlet}(\alpha)$; we draw converter
functions and their hyper-parameters from richer families; and, for
regression, we disable the post-hoc monotone warping of generated
targets, preserving the native smoothness of the causal generating
mechanism (Section~\ref{sec:prior}).

We introduce \model{}, built on TabICLv2: our additions are
deliberately small, and all of them act upstream of the ICL stack.
On the encoder
side, we introduce a two-pass column--row stack that deepens
cross-feature interaction ahead of row compression, adopting the
learned Fourier scalar encoding established by TabFM and Causilo
(Section~\ref{sec:arch}).
Across the two benchmark suites (Table~\ref{tab:benchmarks}), \model{}
builds on the TabICLv2 baseline with consistent gains. Its strongest and most consistent results are observed
on regression: \model{} achieves a 10.1 average per-dataset rank across
the benchmark's full 13-dataset regression suite, outperforming TabFM
(10.8; \citealp{kong2026tabfm}), Mitra-v2 (11.2), EXAONE-Tabular
(14.5; \citealp{eo2026exaone}), and TabPFN-3 (18.3; \citealp{tabpfn3}).
Beyond regression, \model{} remains
competitive across mixed-task evaluations, ranking 11th overall
(Section~\ref{sec:eval}).

\section{Related Work}
\label{sec:related}
\textbf{Synthetic-prior tabular foundation models.}
Prior-fitted networks established that a transformer pretrained on
synthetic datasets approximates Bayesian in-context prediction in a
single forward pass \citep{muller2022pfns}, with in-context learning
theory following for simple function classes
\citep{garg2022transformers}. TabPFN brought the paradigm to tabular
data, pretraining on synthetic datasets sampled from hand-designed
structural causal models for small classification tables
\citep{hollmann2023tabpfnpfn}; TabPFN v2 broadened the prior family
and extended coverage to regression at scale with a cell-based
alternating row/column attention \citep{hollmann2025tabpfnnature}; and
TabICL adopted an SCM-based prior pipeline together with the
column-then-row design and row compression for linear-time ICL
\citep{qu2025tabicl,qu2026tabiclv2}. TabFM combines alternating
attention with row compression and ICL \citep{kong2026tabfm}, and
Causilo inserts an additional row-attention module before compression
\citep{nums2026causilo}. \model{}
differs from all of these by targeting the \emph{pretraining corpus}
rather than the attention pattern: we modify the prior at two specific
points, and our two-pass encoding stays within the TabICLv2
framework so that row compression and TabICLv2's efficient inference
machinery are preserved.

\textbf{Evolution of encoder designs.} The encoder side has evolved
through several regimes. TabPFN v1 lets individual cells interact
across both rows and columns with per-cell attention---expressive,
but quadratic in the number of cells and tightly bounded in context
length---and TabPFN v2 extends and scales the same cell-based
row--column interaction at substantial compute cost
\citep{hollmann2023tabpfnpfn,hollmann2025tabpfnnature}. TabDPT takes
the opposite, efficiency-first route: each entire row is compressed
into a single embedding, and attention runs over row tokens only
\citep{ma2024tabdpt}. TabICL introduced the now-common middle
ground---three column-attention layers followed by three
row-attention layers, then \emph{row compression} into compact
embeddings on which the ICL transformer operates---trading a bounded
amount of pre-compression interaction for linear-time inference
\citep{qu2025tabicl,qu2026tabiclv2}. Two lines extend this pattern
by iterating the column-then-row pass \emph{twice}. TabFM
\citep{kong2026tabfm} keeps the full feature grid throughout two
untied column/row stages and pools rows only at the end, at roughly
double TabICL's parameters; Causilo \citep{nums2026causilo} does so
at TabICL's own budget, with a Set-Transformer-style aggregation in
place of CLS pooling. Causilo matches TabFM on the leaderboards,
suggesting that added parameters pay off better in the
pre-compression encoder than inside the ICL stack. On the capacity
side, Xiaomi-TabLDM extends a TabICL-like backbone's ICL attention
to 24 layers, the last eight mixture-of-experts, at roughly double
the parameter count \citep{xiaomi2026tabldm}, and TabPFN-3 extends
its ICL transformer to 24 layers at a similarly doubled budget
\citep{tabpfn3}. Kumo goes
further still, iterating column--row interaction repeatedly
\emph{after} compression. \model{} follows the same recipe at
TabICLv2's own parameter budget---an additional column-then-row
stage before row compression, with a learned Fourier scalar
embedding at the front---and keeps the
post-compression ICL stack unchanged.

\textbf{Real-data pretraining and prior design.}
TabDPT shows that large-scale pretraining on real tables can be
competitive with synthetic priors \citep{ma2024tabdpt}, and Real-TabPFN
improves TabPFN through continued pretraining on real
data~\citep{realtabpfn25}. On the
synthetic side, prior work has focused on prior \emph{expressivity}---richer
SCM families, flexible structural priors, and controllable
difficulty~\citep{hollmann2023tabpfnpfn,qu2025tabicl,xiaomi2026tabldm}---while
TabICLv2 introduces graph- and dataset-level predictability filtering
and studies its effect on pretraining convergence
\citep{qu2026tabiclv2}. \model{} instead
targets the \emph{differential} effect of prior quality control across
task types---which of these aspects affect regression but not
classification.

\textbf{Scalar encodings.}
Learned sine--cosine (Fourier) encodings of scalar values---multi-scale,
bounded feature maps at learned frequencies---were adapted for
tabular foundation models by TabFM and Causilo (sine--cosine encodings
themselves go back to Transformer position embeddings)
\citep{kong2026tabfm,nums2026causilo}. The encoding design itself follows
prior work and is not claimed as novel; likewise, the
two-tier predictability filters are TabICLv2's own. Our new
contributions are the prior-side changes detailed in
Section~\ref{sec:prior}, together with the two-pass column--row stack.

\section{Overall Framework}
\model{} takes a tabular dataset as input and predicts test targets by
conditioning on labeled training examples as in-context demonstrations,
without task-specific training. Its design centers on two components: a
improved synthetic pretraining prior, and an extended architecture that
combines Fourier scalar encoding with two-pass column--row encoding.

\subsection{Architecture}
\label{sec:arch}
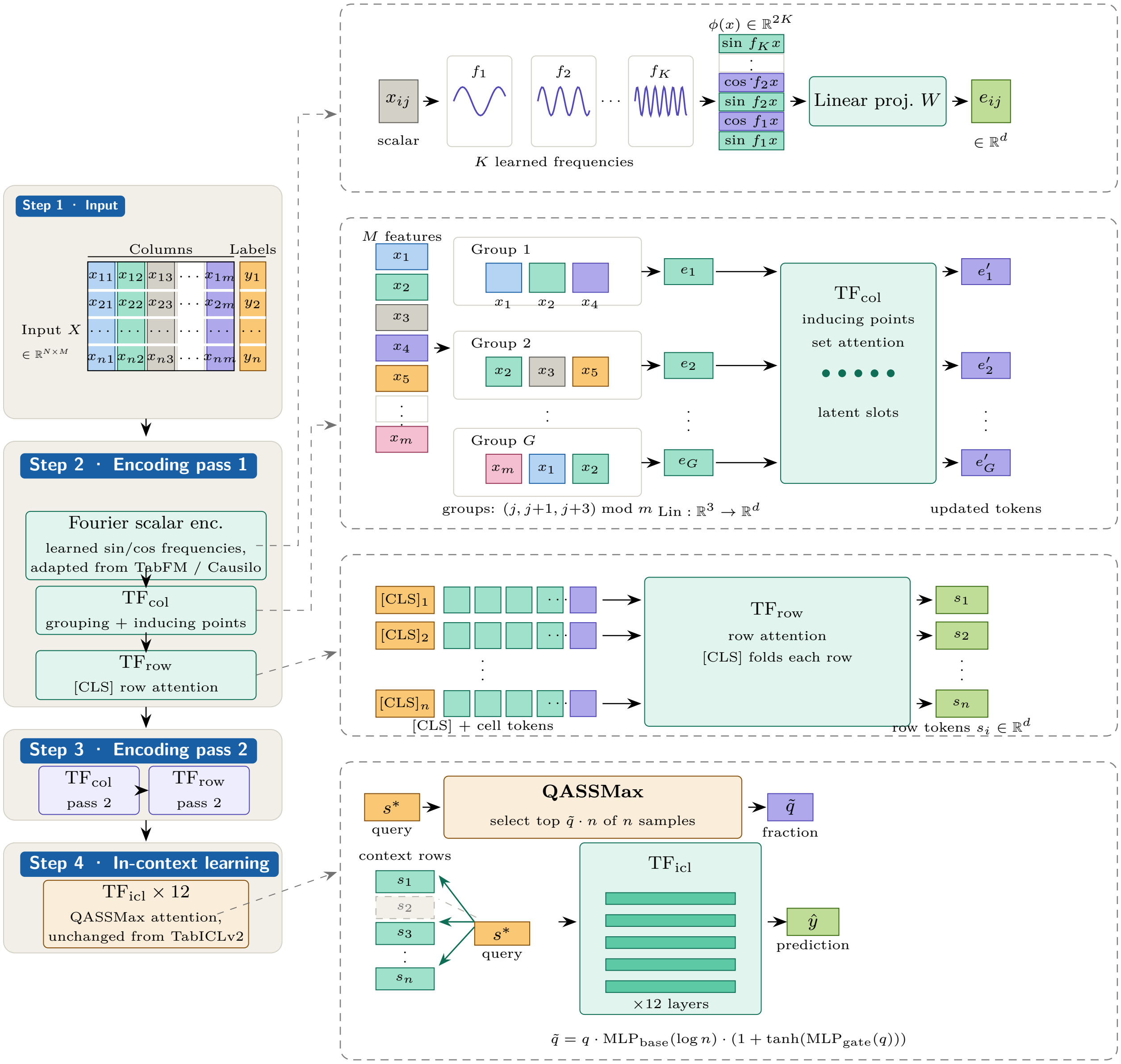
\begin{figure}[H]
\centering
\resizebox{\textwidth}{!}{%
\begin{tikzpicture}[x=1cm,y=1cm,
  cell/.style={inner sep=0.5pt, font=\tiny},
  badge/.style={fill=badgeBlue, text=white, font=\sffamily\scriptsize\bfseries,
    rounded corners=2pt, inner xsep=4pt, inner ysep=2pt},
  panel/.style={draw=panS, fill=panF, rounded corners=5pt, line width=0.4pt},
  zpan/.style={draw=black!50, rounded corners=5pt, line width=0.5pt,
    dashed, dash pattern=on 3.5pt off 2.5pt},
  hbox/.style={draw=tabTealS, fill=tealBg, rounded corners=2pt, line width=0.4pt,
    align=center, font=\scriptsize, inner sep=2pt},
  pbox/.style={draw=tabPurpS, fill=purpBg, rounded corners=2pt, line width=0.4pt,
    align=center, font=\scriptsize, inner sep=2pt},
  abox/.style={draw=tabAmberS, fill=amberBg, rounded corners=2pt, line width=0.4pt,
    align=center, font=\scriptsize, inner sep=2pt},
  arr/.style={-{Stealth[length=2.2mm]}, line width=0.5pt},
  tarr/.style={tabTealS, -{Stealth[length=2mm]}, line width=0.6pt},
  ldr/.style={black!55, -{Stealth[length=2mm]}, dashed, line width=0.4pt}]

\path[panel] (0,9.9) rectangle (4.3,13.5);
\node[badge,anchor=west,font=\sffamily\tiny\bfseries,inner xsep=3pt]
  at (0.18,13.18) {Step 1 \textperiodcentered\ Input};
\node[anchor=west,font=\tiny] at (0.14,11.26) {Input $X$};
\node[anchor=west,font=\fontsize{4.5}{5.5}\selectfont]
  at (0.14,10.90) {$\in\mathbb{R}^{N\times M}$};
\node[font=\tiny] at (2.43,12.52) {Columns};
\draw[line width=0.4pt] (1.3,12.40) -- (3.56,12.40);
\path[fill=tabBlueF,draw=tabBlueS,line width=0.4pt] (1.3,10.64) rectangle +(0.42,1.68);
\path[fill=tabTealF,draw=tabTealS,line width=0.4pt] (1.76,10.64) rectangle +(0.42,1.68);
\path[fill=tabGrayF,draw=tabGrayS,line width=0.4pt] (2.22,10.64) rectangle +(0.42,1.68);
\path[fill=white,draw=panS,line width=0.4pt]   (2.68,10.64) rectangle +(0.42,1.68);
\path[fill=tabPurpF,draw=tabPurpS,line width=0.4pt] (3.14,10.64) rectangle +(0.42,1.68);
\draw[white,line width=1.2pt] (1.3,11.04) -- (3.56,11.04);
\draw[white,line width=1.2pt] (1.3,11.46) -- (3.56,11.46);
\draw[white,line width=1.2pt] (1.3,11.88) -- (3.56,11.88);
\draw[line width=0.4pt] (1.3,10.64) rectangle (3.56,12.32);
\node[cell] at (1.51,12.09) {$x_{11}$}; \node[cell] at (1.97,12.09) {$x_{12}$};
\node[cell] at (2.43,12.09) {$x_{13}$}; \node[cell] at (2.89,12.09) {$\cdots$};
\node[cell] at (3.35,12.09) {$x_{1m}$};
\node[cell] at (1.51,11.67) {$x_{21}$}; \node[cell] at (1.97,11.67) {$x_{22}$};
\node[cell] at (2.43,11.67) {$x_{23}$}; \node[cell] at (2.89,11.67) {$\cdots$};
\node[cell] at (3.35,11.67) {$x_{2m}$};
\node[cell] at (1.51,11.25) {$\cdots$};  \node[cell] at (1.97,11.25) {$\cdots$};
\node[cell] at (2.43,11.25) {$\cdots$};  \node[cell] at (2.89,11.25) {$\cdots$};
\node[cell] at (3.35,11.25) {$\cdots$};
\node[cell] at (1.51,10.83) {$x_{n1}$}; \node[cell] at (1.97,10.83) {$x_{n2}$};
\node[cell] at (2.43,10.83) {$x_{n3}$}; \node[cell] at (2.89,10.83) {$\cdots$};
\node[cell] at (3.35,10.83) {$x_{nm}$};
\path[fill=tabAmberF,draw=tabAmberS,line width=0.4pt] (3.65,10.64) rectangle +(0.4,1.68);
\draw[white,line width=1.2pt] (3.65,11.04) -- (4.05,11.04);
\draw[white,line width=1.2pt] (3.65,11.46) -- (4.05,11.46);
\draw[white,line width=1.2pt] (3.65,11.88) -- (4.05,11.88);
\node[cell] at (3.85,12.09) {$y_1$}; \node[cell] at (3.85,11.67) {$y_2$};
\node[cell] at (3.85,11.25) {$\cdots$}; \node[cell] at (3.85,10.83) {$y_n$};
\node[font=\tiny] at (3.85,12.52) {Labels};
\draw[line width=0.4pt] (3.65,12.40) -- (4.05,12.40);

\path[panel] (0,5.45) rectangle (4.3,9.55);
\node[badge,anchor=west] at (0.25,9.17) {Step 2 \textperiodcentered\ Encoding pass 1};
\node[hbox,minimum width=3.4cm,minimum height=0.68cm]
  at (2.2,7.94) {Fourier scalar enc.\\[1pt]{\tiny learned sin/cos frequencies,}\\[-1pt]{\tiny adapted from TabFM / Causilo}};
\node[hbox,minimum width=3.4cm,minimum height=0.68cm]
  at (2.2,6.94) {$\mathrm{TF}_{\mathrm{col}}$\\[1pt]{\tiny grouping + inducing points}};
\node[hbox,minimum width=3.4cm,minimum height=0.68cm]
  at (2.2,5.94) {$\mathrm{TF}_{\mathrm{row}}$\\[1pt]{\tiny $[\mathrm{CLS}]$ row attention}};
\draw[arr] (2.2,7.6) -- (2.2,7.28);
\draw[arr] (2.2,6.6) -- (2.2,6.28);

\path[panel] (0,3.7) rectangle (4.3,5.1);
\node[badge,anchor=west] at (0.25,4.77) {Step 3 \textperiodcentered\ Encoding pass 2};
\node[pbox,minimum width=1.55cm,minimum height=0.62cm]
  at (1.32,4.16) {$\mathrm{TF}_{\mathrm{col}}$\\[1pt]{\tiny pass 2}};
\node[pbox,minimum width=1.55cm,minimum height=0.62cm]
  at (3.02,4.16) {$\mathrm{TF}_{\mathrm{row}}$\\[1pt]{\tiny pass 2}};
\draw[arr] (2.1,4.16) -- (2.24,4.16);

\path[panel] (0,1.65) rectangle (4.3,3.35);
\node[badge,anchor=west] at (0.25,3.02) {Step 4 \textperiodcentered\ In-context learning};
\node[abox,minimum width=2.2cm,minimum height=0.7cm]
  at (2.2,2.25) {$\mathrm{TF}_{\mathrm{icl}}\times 12$\\[1pt]{\tiny QASSMax attention,}\\[-1pt]{\tiny unchanged from TabICLv2}};

\draw[arr] (2.2,9.9) -- (2.2,9.62);
\draw[arr] (2.2,5.45) -- (2.2,5.12);
\draw[arr] (2.2,3.7) -- (2.2,3.37);

\draw[ldr] (3.9,7.94) -- (4.55,7.94) -- (4.55,14.6) -- (5.15,14.6);
\draw[ldr] (3.9,6.94) -- (4.75,6.94) -- (4.75,9.8) -- (5.15,9.8);
\draw[ldr] (3.9,5.94) -- (5.15,6.3);
\draw[ldr] (3.3,2.25) -- (5.15,2.9);

\path[zpan] (5.2,13.4) rectangle (17.2,16.3);
\node[cell,fill=tabGrayF,draw=tabGrayS,minimum width=0.6cm,minimum height=0.66cm,
  font=\scriptsize] (xij) at (6.1,14.8) {$x_{ij}$};
\node[font=\tiny] at (6.1,14.2) {scalar};
\draw[arr] (6.48,14.8) -- (6.73,14.8);
\draw[fill=white,draw=panS,line width=0.4pt,rounded corners=2pt] (6.85,14.1) rectangle +(1.0,1.4);
\node[font=\tiny] at (7.35,15.26) {$f_1$};
\draw[tabPurpS,line width=0.7pt,domain=6.95:7.75,samples=30]
  plot (\x,{14.8+0.22*sin(650*(\x-6.95))});
\draw[fill=white,draw=panS,line width=0.4pt,rounded corners=2pt] (8.15,14.1) rectangle +(1.0,1.4);
\node[font=\tiny] at (8.65,15.26) {$f_2$};
\draw[tabPurpS,line width=0.7pt,domain=8.25:9.05,samples=40]
  plot (\x,{14.8+0.22*sin(1300*(\x-8.25))});
\node[font=\tiny] at (9.4,14.8) {$\cdots$};
\draw[fill=white,draw=panS,line width=0.4pt,rounded corners=2pt] (9.65,14.1) rectangle +(1.0,1.4);
\node[font=\tiny] at (10.15,15.26) {$f_K$};
\draw[tabPurpS,line width=0.7pt,domain=9.75:10.55,samples=60]
  plot (\x,{14.8+0.22*sin(2600*(\x-9.75))});
\node[font=\tiny] at (8.5,13.85) {$K$ learned frequencies};
\draw[arr] (10.75,14.8) -- (11.0,14.8);
\foreach \yy/\lb/\cf/\cs in
  {14.05/{sin $f_1x$}/tabTealF/tabTealS, 14.35/{cos $f_1x$}/tabPurpF/tabPurpS,
   14.65/{sin $f_2x$}/tabTealF/tabTealS, 14.95/{cos $f_2x$}/tabPurpF/tabPurpS,
   15.25/{$\vdots$}/white/panS,         15.55/{sin $f_Kx$}/tabTealF/tabTealS}
  \path[fill=\cf,draw=\cs,line width=0.4pt] (11.05,\yy) rectangle +(1.0,0.28)
    node[cell,midway] {\lb};
\node[font=\tiny] at (11.55,16.0) {$\phi(x)\in\mathbb{R}^{2K}$};
\draw[arr] (12.15,14.8) -- (12.4,14.8);
\node[hbox,minimum width=2.1cm,minimum height=0.76cm]
  at (13.5,14.8) {Linear proj.\ $W$};
\draw[arr] (14.65,14.8) -- (14.9,14.8);
\node[cell,fill=tabGreenF,draw=tabGreenS,minimum width=0.6cm,minimum height=0.66cm,
  font=\scriptsize] at (15.25,14.8) {$e_{ij}$};
\node[font=\tiny] at (15.25,14.2) {$\in\mathbb{R}^{d}$};

\path[zpan] (5.2,8.2) rectangle (17.2,13.0);
\node[font=\tiny] at (6.15,12.72) {$M$ features};
\foreach \yy/\lb/\cf/\cs in
  {12.2/{$x_1$}/tabBlueF/tabBlueS, 11.73/{$x_2$}/tabTealF/tabTealS,
   11.26/{$x_3$}/tabGrayF/tabGrayS, 10.79/{$x_4$}/tabPurpF/tabPurpS,
   10.32/{$x_5$}/tabAmberF/tabAmberS, 9.85/{$\vdots$}/white/panS,
   9.38/{$x_m$}/tabPinkF/tabPinkS}
  \path[fill=\cf,draw=\cs,line width=0.4pt] (5.75,\yy) rectangle +(0.8,0.4)
    node[cell,midway] {\lb};
\draw[arr] (6.65,10.99) -- (6.9,10.99);
\draw[fill=white,draw=panS,line width=0.4pt,rounded corners=2pt] (6.95,11.65) rectangle +(2.9,1.05);
\node[font=\tiny,anchor=west] at (7.08,12.5) {Group 1};
\path[fill=tabBlueF,draw=tabBlueS,line width=0.4pt] (7.45,11.85) rectangle +(0.55,0.45);
\path[fill=tabTealF,draw=tabTealS,line width=0.4pt] (8.12,11.85) rectangle +(0.55,0.45);
\path[fill=tabPurpF,draw=tabPurpS,line width=0.4pt] (8.79,11.85) rectangle +(0.55,0.45);
\node[cell] at (7.72,11.67) {$x_1$}; \node[cell] at (8.39,11.67) {$x_2$};
\node[cell] at (9.06,11.67) {$x_4$};
\draw[fill=white,draw=panS,line width=0.4pt,rounded corners=2pt] (6.95,10.2) rectangle +(2.9,1.05);
\node[font=\tiny,anchor=west] at (7.08,11.05) {Group 2};
\path[fill=tabTealF,draw=tabTealS,line width=0.4pt] (7.45,10.4) rectangle +(0.55,0.45);
\path[fill=tabGrayF,draw=tabGrayS,line width=0.4pt] (8.12,10.4) rectangle +(0.55,0.45);
\path[fill=tabAmberF,draw=tabAmberS,line width=0.4pt] (8.79,10.4) rectangle +(0.55,0.45);
\node[cell] at (7.72,10.62) {$x_2$}; \node[cell] at (8.39,10.62) {$x_3$};
\node[cell] at (9.06,10.62) {$x_5$};
\node[font=\tiny] at (8.4,9.97) {$\vdots$};
\draw[fill=white,draw=panS,line width=0.4pt,rounded corners=2pt] (6.95,8.7) rectangle +(2.9,1.05);
\node[font=\tiny,anchor=west] at (7.08,9.55) {Group $G$};
\path[fill=tabPinkF,draw=tabPinkS,line width=0.4pt] (7.45,8.9) rectangle +(0.55,0.45);
\path[fill=tabBlueF,draw=tabBlueS,line width=0.4pt] (8.12,8.9) rectangle +(0.55,0.45);
\path[fill=tabTealF,draw=tabTealS,line width=0.4pt] (8.79,8.9) rectangle +(0.55,0.45);
\node[cell] at (7.72,9.12) {$x_m$}; \node[cell] at (8.39,9.12) {$x_1$};
\node[cell] at (9.06,9.12) {$x_2$};
\draw[arr] (9.85,12.17) -- (10.15,12.17);
\draw[arr] (9.85,10.72) -- (10.15,10.72);
\draw[arr] (9.85,9.22) -- (10.15,9.22);
\node[cell,fill=tabTealF,draw=tabTealS,minimum width=0.75cm,minimum height=0.4cm]
  at (10.58,12.17) {$e_1$};
\node[cell,fill=tabTealF,draw=tabTealS,minimum width=0.75cm,minimum height=0.4cm]
  at (10.58,10.72) {$e_2$};
\node[font=\tiny] at (10.58,9.97) {$\vdots$};
\node[cell,fill=tabTealF,draw=tabTealS,minimum width=0.75cm,minimum height=0.4cm]
  at (10.58,9.22) {$e_G$};
\draw[fill=tealBg,draw=tabTealS,line width=0.4pt,rounded corners=2pt] (12.0,8.9) rectangle +(2.4,3.4);
\node[font=\scriptsize\bfseries] at (13.2,11.82) {$\mathrm{TF}_{\mathrm{col}}$};
\node[font=\tiny] at (13.2,11.42) {inducing points};
\node[font=\tiny] at (13.2,11.08) {set attention};
\foreach \dx in {0,1,2,3,4}
  \fill[tabTealS] (12.7+0.25*\dx,10.6) circle (0.06);
\node[font=\tiny] at (13.2,10.0) {latent slots};
\draw[arr] (11.0,12.17) -- (11.95,12.17);
\draw[arr] (11.0,10.72) -- (11.95,10.72);
\draw[arr] (11.0,9.22) -- (11.95,9.22);
\draw[arr] (14.5,12.17) -- (14.75,12.17);
\draw[arr] (14.5,10.72) -- (14.75,10.72);
\draw[arr] (14.5,9.22) -- (14.75,9.22);
\node[cell,fill=tabPurpF,draw=tabPurpS,minimum width=0.75cm,minimum height=0.4cm]
  at (15.17,12.17) {$e'_1$};
\node[cell,fill=tabPurpF,draw=tabPurpS,minimum width=0.75cm,minimum height=0.4cm]
  at (15.17,10.72) {$e'_2$};
\node[font=\tiny] at (15.17,9.97) {$\vdots$};
\node[cell,fill=tabPurpF,draw=tabPurpS,minimum width=0.75cm,minimum height=0.4cm]
  at (15.17,9.22) {$e'_G$};
\node[font=\tiny] at (8.4,8.5) {groups: $(j,j{+}1,j{+}3)\bmod m$};
\node[font=\tiny] at (10.9,8.5) {$\mathrm{Lin}:\mathbb{R}^3\to\mathbb{R}^d$};
\node[font=\tiny] at (15.17,8.5) {updated tokens};

\path[zpan] (5.2,5.0) rectangle (17.2,7.8);
\foreach \yy/\lb in {7.1/{$[\mathrm{CLS}]_1$}, 6.55/{$[\mathrm{CLS}]_2$}, 5.5/{$[\mathrm{CLS}]_n$}}
  \node[cell,fill=tabAmberF,draw=tabAmberS,minimum width=0.9cm,minimum height=0.42cm]
    at (6.2,\yy) {\lb};
\foreach \yy in {7.1,6.55,5.5}
  \foreach \xx in {6.8,7.28,7.76,8.24}
    \path[fill=tabTealF,draw=tabTealS,line width=0.4pt] (\xx,\yy-0.2) rectangle +(0.4,0.4);
\foreach \yy in {7.1,6.55,5.5}
  \node[font=\tiny] at (8.58,\yy) {$\cdots$};
\foreach \yy in {7.1,6.55,5.5}
  \path[fill=tabPurpF,draw=tabPurpS,line width=0.4pt] (8.75,\yy-0.2) rectangle +(0.4,0.4);
\node[font=\tiny] at (7.4,6.12) {$\vdots$};
\node[font=\tiny] at (7.4,5.15) {$[\mathrm{CLS}]$ + cell tokens};
\draw[arr] (9.25,7.1) -- (9.85,7.1);
\draw[arr] (9.25,6.55) -- (9.85,6.55);
\draw[arr] (9.25,5.5) -- (9.85,5.5);
\draw[fill=tealBg,draw=tabTealS,line width=0.4pt,rounded corners=2pt] (9.9,5.15) rectangle +(4.1,2.3);
\node[font=\scriptsize\bfseries] at (11.95,6.95) {$\mathrm{TF}_{\mathrm{row}}$};
\node[font=\tiny] at (11.95,6.55) {row attention};
\node[font=\tiny] at (11.95,6.20) {$[\mathrm{CLS}]$ folds each row};
\draw[arr] (14.1,7.1) -- (14.35,7.1);
\draw[arr] (14.1,6.55) -- (14.35,6.55);
\draw[arr] (14.1,5.5) -- (14.35,5.5);
\node[cell,fill=tabGreenF,draw=tabGreenS,minimum width=0.8cm,minimum height=0.42cm]
  at (14.8,7.1) {$s_1$};
\node[cell,fill=tabGreenF,draw=tabGreenS,minimum width=0.8cm,minimum height=0.42cm]
  at (14.8,6.55) {$s_2$};
\node[font=\tiny] at (14.8,6.12) {$\vdots$};
\node[cell,fill=tabGreenF,draw=tabGreenS,minimum width=0.8cm,minimum height=0.42cm]
  at (14.8,5.5) {$s_n$};
\node[font=\tiny] at (14.8,5.15) {row tokens $s_i\in\mathbb{R}^d$};

\path[zpan] (5.2,0) rectangle (17.2,4.6);
\node[cell,fill=tabAmberF,draw=tabAmberS,minimum width=0.85cm,minimum height=0.42cm,
  font=\scriptsize] at (6.0,3.9) {$s^{*}$};
\node[font=\tiny] at (6.0,3.52) {query};
\draw[arr] (6.47,3.9) -- (6.75,3.9);
\draw[fill=amberBg,draw=tabAmberS,line width=0.4pt,rounded corners=2pt] (6.8,3.42) rectangle +(4.6,0.96);
\node[font=\scriptsize\bfseries] at (9.1,4.12) {QASSMax};
\node[font=\tiny] at (9.1,3.68) {select top $\tilde q\cdot n$ of $n$ samples};
\draw[arr] (11.5,3.9) -- (11.78,3.9);
\node[cell,fill=tabPurpF,draw=tabPurpS,minimum width=0.7cm,minimum height=0.42cm,
  font=\scriptsize] at (12.15,3.9) {$\tilde q$};
\node[font=\tiny] at (12.15,3.52) {fraction};
\node[font=\tiny] at (6.2,3.15) {context rows};
\node[cell,fill=tabTealF,draw=tabTealS,minimum width=0.9cm,minimum height=0.34cm]
  at (6.2,2.75) {$s_1$};
\node[cell,fill=panF,draw=fadedS,dashed,minimum width=0.9cm,minimum height=0.34cm,
  text=black!55] at (6.2,2.35) {$s_2$};
\node[cell,fill=tabTealF,draw=tabTealS,minimum width=0.9cm,minimum height=0.34cm]
  at (6.2,1.95) {$s_3$};
\node[font=\tiny] at (6.2,1.62) {$\vdots$};
\node[cell,fill=tabTealF,draw=tabTealS,minimum width=0.9cm,minimum height=0.34cm]
  at (6.2,1.25) {$s_n$};
\node[cell,fill=tabAmberF,draw=tabAmberS,minimum width=0.85cm,minimum height=0.36cm,
  font=\scriptsize] at (7.7,1.95) {$s^{*}$};
\node[font=\tiny] at (7.7,1.6) {query};
\draw[tarr] (7.27,2.13) -- (6.72,2.85);
\draw[tarr] (7.27,2.13) -- (6.72,2.13);
\draw[tarr] (7.27,2.05) -- (6.72,1.45);
\draw[fadedS,dashed,line width=0.4pt] (7.3,2.2) -- (6.72,2.48);
\draw[arr] (8.57,2.13) -- (8.85,2.13);
\draw[fill=tealBg,draw=tabTealS,line width=0.4pt,rounded corners=2pt] (8.9,0.7) rectangle +(2.8,2.65);
\node[font=\scriptsize\bfseries] at (10.3,3.02) {$\mathrm{TF}_{\mathrm{icl}}$};
\foreach \yy in {2.4,2.06,1.72,1.38,1.04}
  \path[fill=tealMid,draw=tabTealS,line width=0.4pt] (9.3,\yy) rectangle +(2.0,0.18);
\node[font=\tiny] at (10.3,0.85) {$\times 12$ layers};
\draw[arr] (11.8,2.13) -- (12.1,2.13);
\node[cell,fill=tabGreenF,draw=tabGreenS,minimum width=0.8cm,minimum height=0.42cm,
  font=\scriptsize] at (12.5,2.13) {$\hat{y}$};
\node[font=\tiny] at (12.5,1.76) {prediction};
\node[font=\tiny] at (11.2,0.3)
  {$\tilde q = q\cdot\mathrm{MLP}_{\mathrm{base}}(\log n)\cdot
    (1+\tanh(\mathrm{MLP}_{\mathrm{gate}}(q)))$};
\end{tikzpicture}%
}
\caption{Overall architecture of \model{}. \textbf{Left:} the main
pipeline. The input matrix $X\in\mathbb{R}^{N\times M}$ with targets
$y_i\in\mathcal{Y}$ is encoded in two passes (Steps~2--3) and processed
by a 12-layer in-context transformer (Step~4). \textbf{Right (dashed
zoom-ins, top to bottom):} Fourier scalar encoding with $K$ learned
frequencies; repeated feature grouping with circular shifts
$(j,j{+}1,j{+}3)\bmod m$ and inducing-point set attention
($\mathrm{TF}_{\mathrm{col}}$); $[\mathrm{CLS}]$-based row folding
($\mathrm{TF}_{\mathrm{row}}$); and query-adaptive sample selection
(QASSMax) inside $\mathrm{TF}_{\mathrm{icl}}$, where the query row
$s^{*}$ attends only to the top $\tilde q\cdot n$ context rows.}
\label{fig:arch}
\end{figure}

The architecture of \model{} is illustrated in Figure~\ref{fig:arch}.
Following TabICLv2 \citep{qu2026tabiclv2}, \model{} chains column-wise
embedding, row-wise interaction, and dataset-wise in-context learning,
preserving the efficiency of TabICL with a runtime complexity of
$O(n^2 + nm^2)$ for tables with $n$ rows and $m$ columns.

\textbf{The TabICLv2 framework \citep{qu2026tabiclv2}.} Given an input
table $X \in \mathbb{R}^{n \times m}$, the framework proceeds as follows.

\begin{itemize}
  \item \textbf{Repeated feature embedding.} Each feature is embedded into
        multiple groups via circular shifts. For a table
        with $m$ columns, the $j$-th group contains the columns at
        positions $(j, j+1, j+3) \bmod m$ (0-based), encoded by a shared
        linear layer $\mathrm{Lin}: \mathbb{R}^3 \rightarrow \mathbb{R}^d$:
        \begin{equation}
        E_1[i,j] = \mathrm{Lin}\big(x_{i,j},\;
        x_{i,(j+1) \bmod m},\; x_{i,(j+3) \bmod m}\big),
        \end{equation}
        and the shift pattern $(0,1,3)$ ensures that no pair of
        columns appears together in more than one group for tables with at
        least 7 columns. This breaks feature symmetries and mitigates
        cross-feature confusion when features share similar distributions,
        at the cost of strict column-permutation invariance---mitigated in
        practice by the random-shuffle ensembles of TabICLv2, which
        \model{} inherits.
  \item \textbf{Target-aware embedding.} Target information is injected
        early: for training samples, a target embedding is added to every
        token,
        $E_2[i,j] = E_1[i,j] + \mathrm{Embed}_{\mathrm{TAE}}(y_i)$,
        allowing the model to determine feature relevance before
        compression.
  \item \textbf{Compression then ICL.} $E_2$ is processed in three
        stages: (1) \emph{column-wise embedding} applies a set transformer
        \citep{lee2019settransformer} $\mathrm{TF}_{\mathrm{col}}$ with
        inducing-point attention to each
        feature group; (2) \emph{row-wise interaction} applies a
        transformer $\mathrm{TF}_{\mathrm{row}}$ with learnable [CLS]
        tokens, whose concatenated states collapse each row into a single
        vector; (3) \emph{dataset-wise ICL} combines the row embeddings
        with target embeddings in a transformer
        $\mathrm{TF}_{\mathrm{icl}}$, where test samples attend to
        training samples to predict $\hat{y}$. Attention in
        $\mathrm{TF}_{\mathrm{icl}}$ and in the inducing-point part of
        $\mathrm{TF}_{\mathrm{col}}$ uses the query-aware scalable softmax
        (QASSMax) of TabICLv2 \citep{qu2026tabiclv2}, which rescales
        queries as a function of $\log n$ to
        mitigate attention fading and improve long-context
        generalization.
  \item \textbf{Quantile regression.} Classification and regression
        models are pretrained separately. The regressor predicts the
        999 quantiles at levels $\alpha_i = i/1000$,
        $i = 1, \dots, 999$, trained with a
        pinball loss summed across quantiles; point predictions average
        the quantiles, and the sorted quantiles reconstruct a
        full predictive distribution from which prediction intervals and
        uncertainty estimates follow.
\end{itemize}

\model{} departs from this framework in two design components:

\begin{itemize}
  \item \textbf{Fourier scalar encoding.} Each observed scalar $x$ is
        first mapped to sine and cosine features at $K=16$ learned
        frequencies,
        $\phi(x) = [\sin(2\pi f_1 x), \cos(2\pi f_1 x), \dots,
        \sin(2\pi f_K x), \cos(2\pi f_K x)]$,
        whose concatenation replaces the raw scalar inside the grouped
        linear projection of repeated feature embedding, following TabFM
        and Causilo \citep{kong2026tabfm,nums2026causilo}. The Fourier
        basis maps every value to a fixed-size, multi-scale, bounded
        representation, letting the model express high-frequency functions
        of its inputs from the very first layer; missing values use a
        separate learned embedding instead.
  \item \textbf{Two-pass column--row encoding.} The column-then-row
        attention pair $(\mathrm{TF}_{\mathrm{col}} \rightarrow
        \mathrm{TF}_{\mathrm{row}})$ is applied
        \emph{twice} before row compression
        and the ICL transformer: the first
        pass builds initial feature interactions
        and row mixtures, and the second pass re-evaluates cross-feature
        dependencies on the updated representations. Both passes run at
        the same hidden width with independent (untied) weights---separate
        parameter blocks rather than a shared stack---following TabFM,
        which alternates two \emph{untied}
        column/row stages and pools rows into fixed-width vectors only
        after both \citep{kong2026tabfm}; the
        folding happens late, so TabICLv2's efficient inference machinery
        is preserved.
\end{itemize}

\noindent The ICL transformer $\mathrm{TF}_{\mathrm{icl}}$ keeps
TabICLv2's 12 layers \citep{qu2026tabiclv2}; since row compression
fixes the per-row representation length, ICL depth contributes
context-processing steps at no asymptotic context cost.

\subsection{Pretraining}
We pretrain classification and regression models separately, following the
TabICLv2 pretraining protocol \citep{qu2026tabiclv2}. The classifier is
trained with a cross-entropy loss and the regressor with the 999-quantile
pinball loss of Section~\ref{sec:arch}. TabICLv2 pretrains in three stages
that progressively expand dataset size, with up to 100 features and a batch
size of 64 throughout: \emph{Stage 1}: 500K steps on datasets with 1{,}024
samples (30--90\% for training), maximum learning rate 8e-4;
\emph{Stage 2}: 40K steps on datasets with 400--10{,}240 samples
(log-uniform, $\sim$80\% for training), 1e-4; \emph{Stage 3}: 10K steps on
datasets with 400--60K samples, 2e-5. Optimization uses the Muon
optimizer \citep{jordan2024muon} (replacing AdamW) with weight decay 0.01,
gradient clipping at 10 for
Stages 1--2, and a cosine learning-rate schedule. We train \model{} with
the improved prior of Section~\ref{sec:prior}: the three stages are
trained on 8$\times$H20 GPUs---around 8--12 days for Stage 1 and 3 days
each for Stages 2 and 3, roughly 2{,}700--3{,}500 GPU-hours in total.
\subsection{Synthetic Prior}
\label{sec:prior}
\begin{figure}[H]
\centering
\begin{tikzpicture}[node distance=0.55cm and 0.45cm,
  box/.style={draw, rounded corners=2pt, align=center, inner sep=5pt,
              minimum height=1.05cm, font=\small},
  drop/.style={draw=red!60, dashed, rounded corners=2pt, align=center,
              inner sep=4pt, font=\scriptsize, text=red!70!black}]
\node[box] (cfg) {1. Task\\ configuration};
\node[box, right=of cfg] (graph) {2. Graph sampling\\ \& propagation};
\node[box, right=of graph] (tab) {3. Tabular\\ realization};
\node[box, right=of tab, draw=green!50!black] (corpus) {Pretraining\\ corpus};
\node[drop, below=0.55cm of graph] (dg) {unpredictable graphs $\times$};
\draw[-{Stealth}] (cfg) -- (graph);
\draw[-{Stealth}] (graph) -- (tab);
\draw[-{Stealth}] (tab) -- (corpus);
\draw[-{Stealth}, red!60, dashed] (graph) -- (dg);
\end{tikzpicture}
\caption{Schematic overview of the improved synthetic data generation
prior. The dataset-level pipeline samples dataset properties, generates a
graph, and constructs the tabular data; graphs whose targets are
unlearnable are rejected and resampled, so only learnable problems enter
the pretraining corpus.}
\label{fig:prior}
\end{figure}
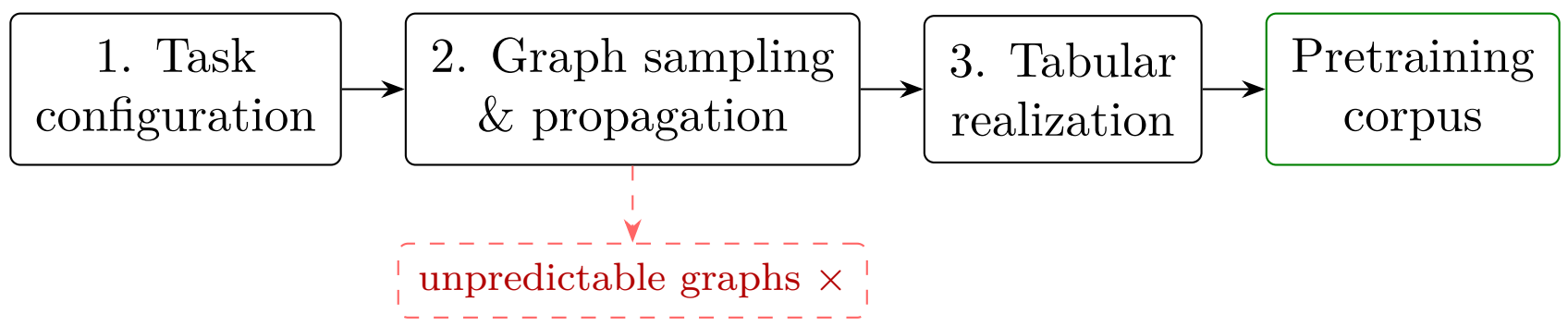

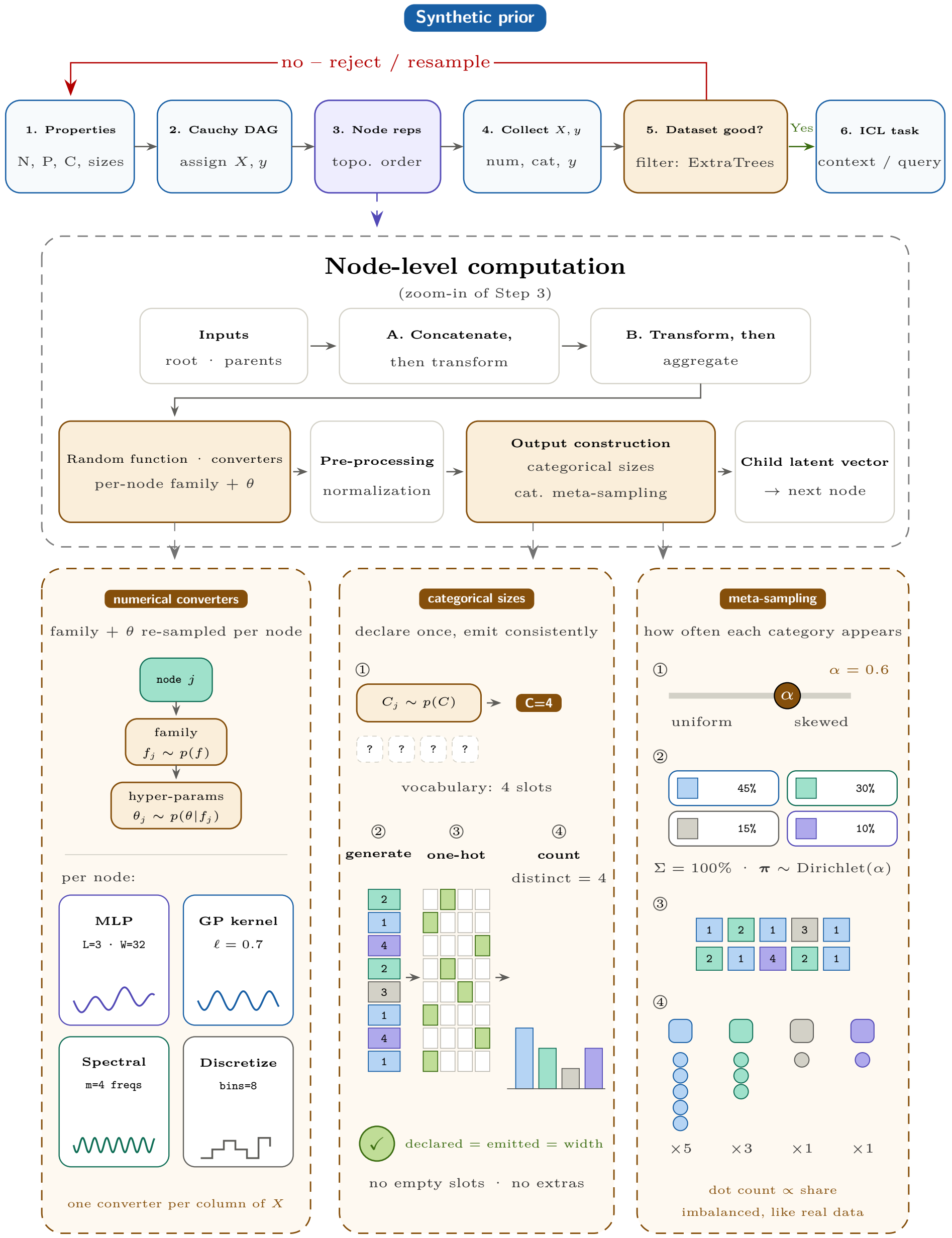
\begin{figure}[H]
\centering
\resizebox{0.75\textwidth}{!}{%
\begin{tikzpicture}[x=0.017241cm,y=-0.017241cm,
  cell/.style={inner sep=0.5pt, font=\fontsize{4}{5}\selectfont},
  txt/.style={inner sep=0.5pt, font=\fontsize{4.5}{5.5}\selectfont, text=black!80},
  hd/.style={inner sep=0.5pt, font=\fontsize{4}{5}\selectfont\bfseries},
  mono/.style={inner sep=0.5pt, font=\fontsize{4}{5}\selectfont\ttfamily},
  topb/.style={fill=badgeBlue, text=white,
    font=\sffamily\bfseries\fontsize{6}{7}\selectfont,
    rounded corners=3.5pt, inner xsep=4pt, inner ysep=2pt},
  fixb/.style={fill=tabAmberS, text=white, font=\sffamily\bfseries\fontsize{4.5}{5.5}\selectfont,
    rounded corners=2pt, inner xsep=3pt, inner ysep=1.5pt},
  tarr/.style={tabGrayS, -{Stealth[length=1.3mm]}, line width=0.45pt},
  ldr/.style={black!55, dashed, -{Stealth[length=1.8mm]}, line width=0.35pt},
  stepb/.style={draw=tabBlueS, fill=stepBg, rounded corners=5pt, line width=0.45pt},
  whiteb/.style={draw=panS, fill=white, rounded corners=4pt, line width=0.4pt},
  amb/.style={draw=tabAmberS, fill=amberBg, rounded corners=4pt, line width=0.5pt},
  fixpan/.style={draw=tabAmberS, fill=tabAmberF, fill opacity=0.10,
    rounded corners=7pt, dashed, dash pattern=on 3.5pt off 2.5pt, line width=0.5pt},
  zpan/.style={draw=black!50, fill=none, rounded corners=7pt, dashed,
    dash pattern=on 3.5pt off 2.5pt, line width=0.5pt}]
\def\cnum#1#2#3{\draw[tabGrayS,line width=0.35pt] (#1,#2) circle (4.5);
  \node[cell] at (#1,#2) {#3};}

\node[topb] at (340,19)
  {Synthetic prior};
\draw[red!70!black, line width=0.45pt, -{Stealth[length=2mm]}]
  (500,76) -- (500,50) -- (60,50) -- (60,73);
\node[fill=white, inner sep=1pt, font=\tiny, text=red!70!black]
  at (278,50) {no -- reject / resample};

\foreach \bx/\bw/\cx/\ta/\tb in {
  15/89/60/{1. Properties}/{N, P, C, sizes},
  120/93/166/{2. Cauchy DAG}/{assign $X,y$},
  332/95/379/{4. Collect $X,y$}/{num, cat, $y$},
  576/89/620/{6. ICL task}/{context / query}}{%
  \path[stepb] (\bx,76) rectangle +(\bw,64);
  \node[hd, font=\fontsize{3.5}{4.5}\selectfont\bfseries] at (\cx,97) {\ta};
  \node[txt] at (\cx,119) {\tb};
}
\path[draw=tabPurpS, fill=purpBg, rounded corners=5pt, line width=0.5pt]
  (229,76) rectangle +(87,64);
\node[hd, font=\fontsize{3.5}{4.5}\selectfont\bfseries] at (272,97) {3. Node reps};
\node[txt] at (272,119) {topo. order};
\path[draw=tabAmberS, fill=amberBg, rounded corners=5pt, line width=0.5pt]
  (443,76) rectangle +(113,64);
\node[hd, font=\fontsize{3.5}{4.5}\selectfont\bfseries] at (499,97) {5. Dataset good?};
\node[txt] at (499,119) {filter: ExtraTrees};
\foreach \ax/\bx in {104/120, 213/229, 316/332, 427/443}
  \draw[tarr] (\ax,108) -- (\bx,108);
\draw[tarr,tabGreenS] (557,108) -- (575,108);
\node[cell, text=tabGreenS] at (566,95) {Yes};
\draw[tabPurpS, dashed, line width=0.45pt, -{Stealth[length=2mm]}]
  (272,140) -- (272,164);

\draw[zpan] (40,170) rectangle (640,385);
\node[font=\scriptsize\bfseries] at (340,192) {Node-level computation};
\node[txt] at (340,210) {(zoom-in of Step 3)};
\path[whiteb] (108,220) rectangle +(116,52);
\node[hd] at (166,239) {Inputs};
\node[txt] at (166,257) {root \textperiodcentered\ parents};
\path[whiteb] (246,220) rectangle +(152,52);
\node[hd] at (322,239) {A. Concatenate,};
\node[txt] at (322,257) {then transform};
\path[whiteb] (420,220) rectangle +(152,52);
\node[hd] at (496,239) {B. Transform, then};
\node[txt] at (496,257) {aggregate};
\draw[tarr] (226,246) -- (244,246);
\draw[tarr] (400,246) -- (418,246);
\draw[tarr] (496,272) -- (496,282) -- (132,282) -- (132,295);

\path[amb] (52,298) rectangle +(160,70);
\node[cell] at (132,324) {Random function \textperiodcentered\ converters};
\node[txt] at (132,342) {per-node family + $\theta$};
\path[whiteb] (226,298) rectangle +(92,70);
\node[hd] at (272,326) {Pre-processing};
\node[txt] at (272,346) {normalization};
\path[amb] (334,298) rectangle +(172,70);
\node[hd] at (420,314) {Output construction};
\node[txt] at (420,330) {categorical sizes};
\node[txt] at (420,348) {cat. meta-sampling};
\path[whiteb] (520,298) rectangle +(110,70);
\node[hd] at (575,326) {Child latent vector};
\node[txt] at (575,346) {$\rightarrow$ next node};
\draw[tarr] (214,333) -- (224,333);
\draw[tarr] (320,333) -- (332,333);
\draw[tarr] (508,333) -- (518,333);
\foreach \x in {132,380,470} \draw[ldr] (\x,368) -- (\x,394);

\path[fixpan] (40,400) rectangle +(186,460);
\node[fixb] at (133,421) {numerical converters};
\node[txt] at (133,444) {family + $\theta$ re-sampled per node};
\path[fill=tabTealF, draw=tabTealS, rounded corners=4pt, line width=0.4pt]
  (108,462) rectangle +(50,30);
\node[mono] at (133,477) {node $j$};
\draw[tarr] (133,492) -- (133,502);
\path[amb] (98,504) rectangle +(70,32);
\node[cell] at (133,514) {family};
\node[mono] at (133,528) {$f_j \sim p(f)$};
\draw[tarr] (133,536) -- (133,546);
\path[amb] (88,548) rectangle +(90,32);
\node[cell] at (133,558) {hyper-params};
\node[mono] at (133,572) {$\theta_j \sim p(\theta|f_j)$};
\draw[panS, line width=0.3pt] (56,598) -- (210,598);
\node[txt, anchor=west] at (52,614) {per node:};
\path[fill=white, draw=tabPurpS, rounded corners=4pt, line width=0.4pt]
  (52,626) rectangle +(76,90);
\node[hd] at (90,644) {MLP};
\node[mono] at (90,660) {L=3 \textperiodcentered\ W=32};
\draw[tabPurpS, line width=0.6pt, domain=62:118, samples=90]
  plot (\x, {699+6*sin(18*(\x-62))+3.5*sin(7*(\x-62))});
\path[fill=white, draw=tabBlueS, rounded corners=4pt, line width=0.4pt]
  (138,626) rectangle +(76,90);
\node[hd] at (176,644) {GP kernel};
\node[mono] at (176,660) {$\ell=0.7$};
\draw[tabBlueS, line width=0.6pt, domain=148:204, samples=80]
  plot (\x, {699+6.5*sin(20*(\x-148))});
\path[fill=white, draw=tabTealS, rounded corners=4pt, line width=0.4pt]
  (52,724) rectangle +(76,90);
\node[hd] at (90,742) {Spectral};
\node[mono] at (90,758) {m=4 freqs};
\draw[tabTealS, line width=0.6pt, domain=62:118, samples=140]
  plot (\x, {799+5*sin(43*(\x-62))});
\path[fill=white, draw=tabGrayS, rounded corners=4pt, line width=0.4pt]
  (138,724) rectangle +(76,90);
\node[hd] at (176,742) {Discretize};
\node[mono] at (176,758) {bins=8};
\draw[tabGrayS, line width=0.6pt]
  (150,808) -- (158,808) -- (158,802) -- (166,802) -- (166,796) --
  (174,796) -- (174,802) -- (182,802) -- (182,808) -- (190,808) --
  (190,794) -- (198,794) -- (198,800);
\node[cell, text=tabAmberS] at (133,840) {one converter per column of $X$};

\path[fixpan] (246,400) rectangle +(190,460);
\node[fixb] at (341,421) {categorical sizes};
\node[txt] at (341,444) {declare once, emit consistently};
\cnum{262}{470}{1}
\path[amb] (258,480) rectangle +(86,26);
\node[mono] at (301,493) {$C_j \sim p(C)$};
\draw[tarr] (348,493) -- (358,493);
\node[fixb] at (384,493) {C=4};
\foreach \x in {258,280,302,324}
  \path[fill=white, draw=panS, dashed, dash pattern=on 1.5pt off 1.5pt,
    rounded corners=2pt, line width=0.3pt] (\x,516) rectangle +(18,18);
\foreach \x in {267,289,311,333} \node[mono] at (\x,525) {?};
\node[txt] at (341,552) {vocabulary: 4 slots};
\cnum{273}{582}{2}
\node[hd] at (273,598) {generate};
\cnum{327}{582}{3}
\node[hd] at (327,598) {one-hot};
\cnum{398}{582}{4}
\node[hd] at (398,598) {count};
\node[txt] at (398,614) {distinct = 4};
\foreach \y in {638,702,734}{%
  \path[fill=tabBlueF,draw=tabBlueS,line width=0.3pt]
    (266,\y) rectangle +(22,14);
  \node[mono] at (277,\y+7) {1};}
\foreach \y in {622,670}{%
  \path[fill=tabTealF,draw=tabTealS,line width=0.3pt]
    (266,\y) rectangle +(22,14);
  \node[mono] at (277,\y+7) {2};}
\foreach \y in {686}{%
  \path[fill=tabGrayF,draw=tabGrayS,line width=0.3pt]
    (266,\y) rectangle +(22,14);
  \node[mono] at (277,\y+7) {3};}
\foreach \y in {654,718}{%
  \path[fill=tabPurpF,draw=tabPurpS,line width=0.3pt]
    (266,\y) rectangle +(22,14);
  \node[mono] at (277,\y+7) {4};}
\foreach \y in {622,638,654,670,686,702,718,734}
  \foreach \x in {304,316,328,340}
    \path[fill=white, draw=fadedS, line width=0.25pt]
      (\x,\y) rectangle +(10,14);
\foreach \x/\y in {316/622, 304/638, 340/654, 316/670, 328/686,
                   304/702, 340/718, 304/734}
  \path[fill=tabGreenF, draw=tabGreenS, line width=0.3pt]
    (\x,\y) rectangle +(10,14);
\draw[tarr] (292,683) -- (302,683);
\draw[tarr] (354,683) -- (364,683);
\path[fill=tabBlueF, draw=tabBlueS, line width=0.3pt]  (368,718) rectangle +(12,42);
\path[fill=tabTealF, draw=tabTealS, line width=0.3pt]  (384,732) rectangle +(12,28);
\path[fill=tabGrayF, draw=tabGrayS, line width=0.3pt]  (400,746) rectangle +(12,14);
\path[fill=tabPurpF, draw=tabPurpS, line width=0.3pt]  (416,732) rectangle +(12,28);
\draw[tabGrayS, line width=0.3pt] (362,760) -- (430,760);
\draw[fill=tabGreenF, draw=tabGreenS, line width=0.5pt] (272,798) circle (12);
\node[cell, text=tabGreenS, font=\tiny\bfseries] at (272,798) {\checkmark};
\node[cell, text=tabGreenS, anchor=west] at (290,798) {declared = emitted = width};
\node[txt] at (341,826) {no empty slots \textperiodcentered\ no extras};

\path[fixpan] (452,400) rectangle +(188,460);
\node[fixb] at (546,421) {meta-sampling};
\node[txt] at (546,444) {how often each category appears};
\cnum{468}{470}{1}
\node[txt, text=tabAmberS, anchor=east] at (628,470) {$\alpha = 0.6$};
\draw[panS, line width=2pt] (474,488) -- (600,488);
\draw[fill=tabAmberS] (556,488) circle (9);
\node[cell, text=white, font=\tiny\bfseries] at (556,488) {$\alpha$};
\node[txt, anchor=west] at (474,506) {uniform};
\node[txt, anchor=east] at (600,506) {skewed};
\cnum{468}{530}{2}
\foreach \x/\y/\cs in {474/540/tabBlueS, 556/540/tabTealS,
                      474/568/tabGrayS, 556/568/tabPurpS}
  \path[fill=white, draw=\cs, rounded corners=3pt, line width=0.4pt]
    (\x,\y) rectangle +(76,24);
\path[fill=tabBlueF,draw=tabBlueS,line width=0.3pt] (480,545) rectangle +(14,14);
\node[mono] at (528,552) {45\%};
\path[fill=tabTealF,draw=tabTealS,line width=0.3pt] (562,545) rectangle +(14,14);
\node[mono] at (610,552) {30\%};
\path[fill=tabGrayF,draw=tabGrayS,line width=0.3pt] (480,573) rectangle +(14,14);
\node[mono] at (528,580) {15\%};
\path[fill=tabPurpF,draw=tabPurpS,line width=0.3pt] (562,573) rectangle +(14,14);
\node[mono] at (610,580) {10\%};
\node[txt] at (546,608)
  {$\Sigma = 100\%$ \textperiodcentered\ $\boldsymbol{\pi} \sim \mathrm{Dirichlet}(\alpha)$};
\cnum{468}{632}{3}
\foreach \x/\cf/\cs/\v/\y in {
  493/tabBlueF/tabBlueS/1/642, 515/tabTealF/tabTealS/2/642,
  537/tabBlueF/tabBlueS/1/642, 559/tabGrayF/tabGrayS/3/642,
  581/tabBlueF/tabBlueS/1/642, 493/tabTealF/tabTealS/2/662,
  515/tabBlueF/tabBlueS/1/662, 537/tabPurpF/tabPurpS/4/662,
  559/tabTealF/tabTealS/2/662, 581/tabBlueF/tabBlueS/1/662}{%
  \path[fill=\cf, draw=\cs, line width=0.3pt] (\x,\y) rectangle +(18,16);
  \node[mono] at (\x+9,\y+8) {\v};}
\cnum{468}{700}{4}
\path[fill=tabBlueF, draw=tabBlueS, rounded corners=2pt, line width=0.35pt]
  (474,712) rectangle +(16,16);
\path[fill=tabTealF, draw=tabTealS, rounded corners=2pt, line width=0.35pt]
  (516,712) rectangle +(16,16);
\path[fill=tabGrayF, draw=tabGrayS, rounded corners=2pt, line width=0.35pt]
  (558,712) rectangle +(16,16);
\path[fill=tabPurpF, draw=tabPurpS, rounded corners=2pt, line width=0.35pt]
  (600,712) rectangle +(16,16);
\foreach \y in {740,751,762,773,784}
  \draw[fill=tabBlueF, draw=tabBlueS, line width=0.35pt] (482,\y) circle (5);
\foreach \y in {740,751,762}
  \draw[fill=tabTealF, draw=tabTealS, line width=0.35pt] (524,\y) circle (5);
\draw[fill=tabGrayF, draw=tabGrayS, line width=0.35pt] (566,740) circle (5);
\draw[fill=tabPurpF, draw=tabPurpS, line width=0.35pt] (608,740) circle (5);
\node[txt] at (482,802) {$\times$5};
\node[txt] at (524,802) {$\times$3};
\node[txt] at (566,802) {$\times$1};
\node[txt] at (608,802) {$\times$1};
\node[cell, text=tabAmberS] at (546,830) {dot count $\propto$ share};
\node[cell, text=tabAmberS] at (546,846) {imbalanced, like real data};
\end{tikzpicture}}
\caption{The improved synthetic prior. \textbf{Top:} the dataset-level
pipeline; datasets failing the signal check are resampled. \textbf{Middle:} the node-level computation of Step 3.
\textbf{Bottom:} the three generation-time changes---richer converter
function families ($f$ and $\theta$ drawn per node), strict one-hot
width matching (declared, emitted, and one-hot widths agree exactly),
and class frequencies from multiple distributions, chiefly
$\boldsymbol{\pi} \sim \mathrm{Dirichlet}(\alpha)$.}
\label{fig:priorfix}
\end{figure}

Following TabPFN and TabICL
\citep{hollmann2023tabpfnpfn,qu2025tabicl,qu2026tabiclv2}, \model{} is
pretrained entirely on synthetic datasets generated from an SCM prior.
Our prior-side changes act across the generation pipeline.
Figure~\ref{fig:prior} summarizes the pipeline in three stages;
Figure~\ref{fig:priorfix} locates our changes within them.

\begin{enumerate}
  \item \textbf{Task configuration.} Each synthetic task begins with a
        draw of its global configuration \citep{qu2026tabiclv2}: the
        number of rows, the number of features, and their ratio; the mix
        of numerical and categorical columns together with their
        cardinalities; the target type---binary, multiclass, or
        regression, with the number of classes drawn for
        classification---and the context--query partition, which fixes
        how many labeled examples serve as in-context demonstrations and
        how many rows are queried.
  \item \textbf{Graph sampling and propagation.} A directed acyclic graph
        is drawn from TabICLv2's random Cauchy mechanism
        \citep{qu2026tabiclv2}; latent vectors propagate in topological
        order, and each node evaluates a random function of its parents
        from eight classes spanning smooth and non-smooth inductive
        biases. In \model{}, the function family $f$ and its
        hyper-parameters $\theta$ are drawn \emph{per node} from richer
        families, $f \sim p(f)$, $\theta \sim p(\theta \mid f)$, so the
        columns of $X$ realize diverse function shapes instead of
        correlated duplicates.
  \item \textbf{Tabular realization.} Graph nodes are designated as
        observed features or the target; a random subset of each
        node's dimensions is revealed, the rest remaining latent.
        Continuous dimensions feed numerical features and regression
        targets, categorical ones arise through discretization or
        softmax sampling. In \model{}, the declared cardinality
        $C_j$, the emitted categories, and the one-hot width of
        feature $j$ agree exactly---no empty or extra slots---so the
        embedding layer learns stable category semantics. Class
        frequencies are drawn from multiple distributions, most
        prominently
        $\boldsymbol{\pi} \sim \mathrm{Dirichlet}(\alpha)$, reproducing
        the class imbalance of real data. TabICLv2's converters may
        additionally apply a post-hoc monotone warping to targets;
        \model{} disables this step, preserving the native output of
        the causal mechanism, which itself still samples non-smooth
        node functions such as tree ensembles. Datasets failing the
        final signal check are rejected and resampled before entering
        the corpus.
\end{enumerate}

Together, these changes raise the quality and diversity of the synthetic
prior: sampled tasks span diverse nonlinear feature interactions,
carry stable categorical representations, and exhibit the skewed
class frequencies of real data.

\section{Evaluation}
\label{sec:eval}
We evaluate \model{} on two public tabular learning benchmarks,
BeyondArena \citep{beyondarena2026} and TabArena
\citep{erickson2026tabarena}, which together cover a broad range of
real-world classification and regression tasks;
Table~\ref{tab:benchmarks} summarizes their composition.

\textbf{BeyondArena} \citep{beyondarena2026} probes generalization
beyond standard IID splits with 142 datasets (98 classification, 44
regression), stratified along four axes: task type (IID / Grouped /
Temporal), dataset scale (Tiny / Small / Medium / Large),
dimensionality (Low / High), and feature type (Text /
High-Cardinality). Its size distribution---80 datasets below 10K rows,
35 between 10K and 100K, and 27 above 100K---stresses regimes that
IID-only suites underrepresent.

\textbf{TabArena} \citep{erickson2026tabarena} is a living benchmark
that compares tabular foundation models and tuned classical ML
pipelines under a unified evaluation protocol. Its suite contains 51
datasets---38 classification and 13 regression (the benchmark's full
regression suite)---with sizes spanning under 10K rows (29 datasets),
10K--100K (20), and over 100K (2). Methods are ranked by an Elo rating
anchored at default Random Forest $=1000$; the leaderboard pools 99
evaluated entries, including the TabPFN family, TabICLv2, TabFM,
Causilo, Kumo, LimiX-2~\citep{limix2025,limix2m},
TabM~\citep{gorishniy2025tabm}, and tuned/ensembled
gradient-boosted trees, and complements Elo with aggregate per-dataset
views (average and
harmonic rank, oracle improvability, and win rate).

\begin{table}[!htb]
\centering
\caption{Statistics of the benchmark suites considered in our evaluation.
Sizes are counted in number of instances per dataset; the size columns
report the number of datasets in each range.}
\label{tab:benchmarks}
\small
\renewcommand{\arraystretch}{1.15}
\begin{tabular*}{\textwidth}{@{\extracolsep{\fill}} l c c c c c c}
\toprule
Benchmark & \# Datasets & Cls. & Reg. & $<$10K & 10K--100K & $>$100K \\
\midrule
BeyondArena & 142 & 98 & 44 & 80 & 35 & 27 \\
TabArena    & 51  & 38 & 13 & 29 & 20 & 2  \\
\bottomrule
\end{tabular*}
\end{table}

\subsection{Main Results}
\textbf{BeyondArena.} Beyond TabArena, we evaluate \model{} on the
BeyondArena suite \citep{beyondarena2026}: 142 datasets (98
classification, 44 regression) with grouped and temporal splits in
addition to IID ones, stratified along task, scale, dimensionality,
and feature-type axes.

\begin{figure}[H]
\centering
\includegraphics[width=0.85\textwidth]{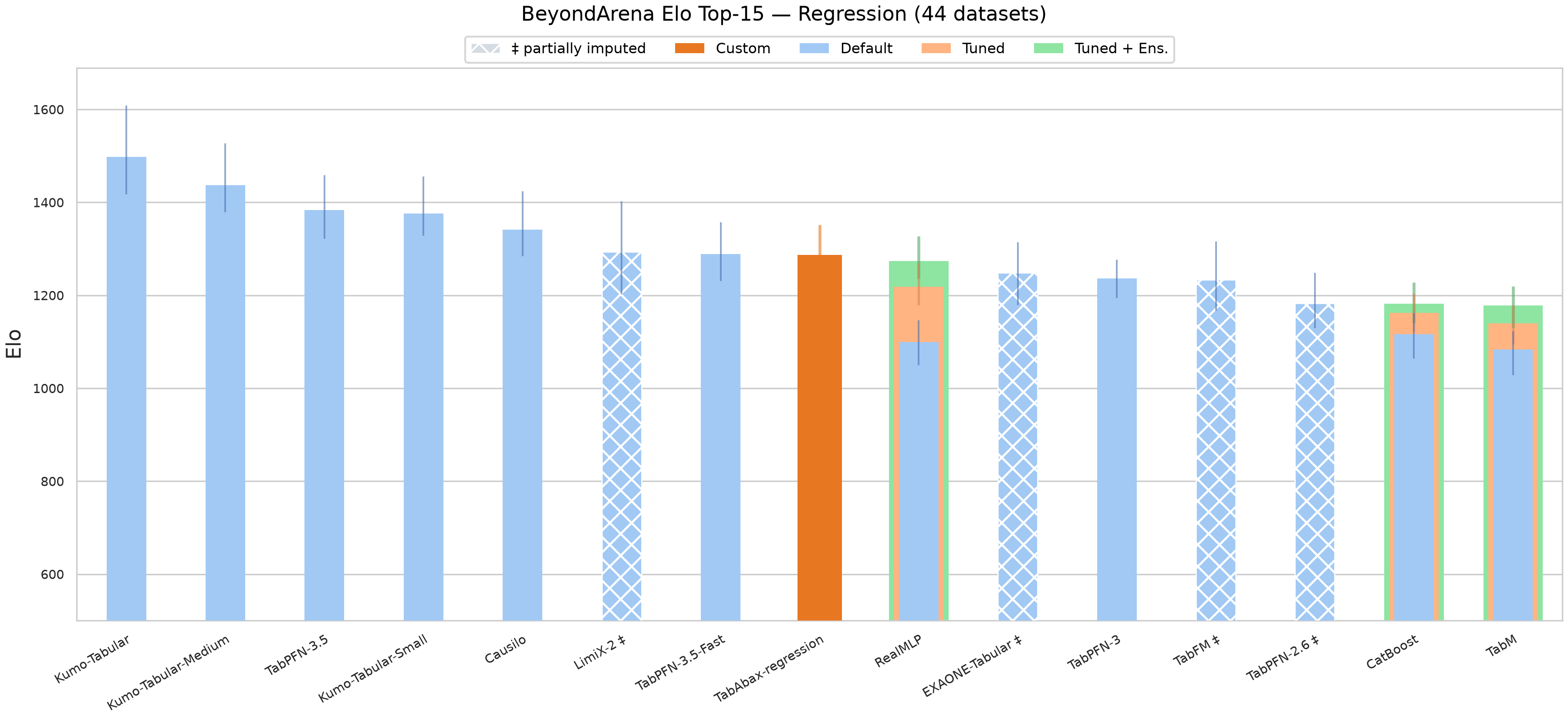}
\caption{BeyondArena regression Elo across the 44 regression datasets
(top-15 leaderboard). \texttt{TabAbax-regression}
(dark orange bar) reaches an Elo rating of 1{,}287, ranking 8th of the 41
evaluated methods, $+$127 over the TabICLv2 baseline (1{,}160), while
the Kumo family, TabPFN-3.5, Causilo, and TabPFN-3.5-Fast
retain their lead. Methods marked $\ddagger$ are partially imputed.}
\label{fig:beyondarena-regression}
\end{figure}

\textbf{Regression.} On the 44 regression datasets of BeyondArena,
\texttt{TabAbax-regression} reaches an Elo rating of
1{,}287, ranking 8th of the 41 evaluated methods---$+$127 over the
TabICLv2 baseline (1{,}160)
(Figure~\ref{fig:beyondarena-regression}; the full 41-method table is
given in Table~\ref{tab:app-beyondarena-reg}).

\begin{figure}[H]
\centering
\includegraphics[width=0.85\textwidth]{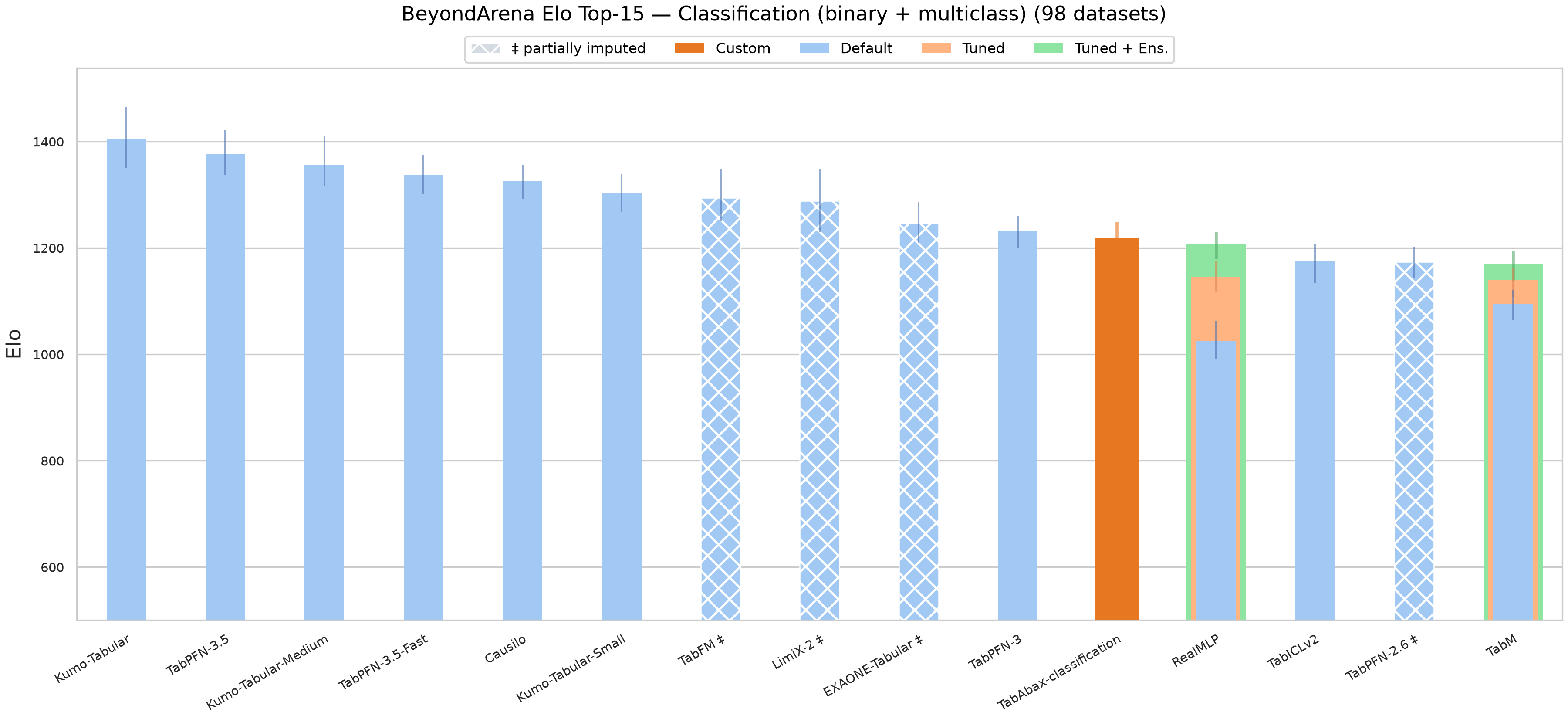}
\caption{BeyondArena classification Elo across the 98 classification
datasets (binary + multiclass; top-15 leaderboard).
\texttt{TabAbax-classification} (dark orange bar) reaches an
Elo rating of 1{,}219, ranking 11th of the 41 evaluated methods, $+$43 over
the TabICLv2 baseline (1{,}176), while the Kumo family, TabPFN-3.5,
TabFM, EXAONE-Tabular, and TabPFN-3 retain their lead.
Methods marked $\ddagger$ are partially imputed.}
\label{fig:beyondarena-classification}
\end{figure}

{\sloppypar
\textbf{Classification.} On the 98 classification datasets of
BeyondArena, \texttt{TabAbax-classification} reaches an Elo rating of
1{,}219, ranking 11th of the 41 evaluated methods---$+$43 over the
TabICLv2 baseline (1{,}176)
(Figure~\ref{fig:beyondarena-classification}; the full 41-method table
is given in Table~\ref{tab:app-beyondarena-cls}).\par}

\textbf{Overall.} On the full 142-dataset suite, \texttt{TabAbax}
reaches an Elo rating of
1{,}238, ranking \textbf{10th} of the 41 evaluated methods---$+$68 over
the TabICLv2 baseline (1{,}170)
(selected methods in Table~\ref{tab:beyond-main}; the full 41-method
table is given in Table~\ref{tab:app-beyondarena}).
The position
is stable across strata: \model{} ranks 9th on Tiny datasets and 10th
or within a few positions of the top ten on every other stratum,
indicating that its performance does not concentrate on a particular
dataset regime---the same healthy-tail behavior observed on TabArena.

\begin{table}[H]
\centering
\caption{BeyondArena leaderboard, selected methods (142 datasets; the
method set follows
Table~\ref{tab:main} for comparability; Rank is the Elo ordinal among
all 41 evaluated methods). Elo$+$ / Elo$-$ are error-bar extents;
Avg.\ rank is the mean per-dataset rank; Improv.\ is the oracle
improvability (lower is better); Winrate is the per-dataset win rate.
The full 41-method table is in Table~\ref{tab:app-beyondarena}.}
\label{tab:beyond-main}
\small
\begin{tabular}{c l r r r r r r}
\toprule
Rank & Method & Elo$\uparrow$ & Elo$+$ & Elo$-$ & Avg.\ rank$\downarrow$ & Improv.\ (\%)$\downarrow$ & Winrate$\uparrow$ \\
\midrule
5  & Causilo                  & 1{,}328.7 & 30.7 & 26.8 & 11.4 & 12.03 & 0.740 \\
8  & TabFM                    & 1{,}274.4 & 39.4 & 36.9 & 13.7 & 10.82 & 0.682 \\
10 & \textbf{\model{} (ours)} & \textbf{1{,}237.7} & \textbf{29.0} & \textbf{29.1} & \textbf{15.4} & \textbf{14.47} & \textbf{0.640} \\
13 & TabPFN-2.6               & 1{,}175.9 & 26.4 & 24.0 & 18.4 & 16.56 & 0.565 \\
16 & TabICLv2 (baseline)      & 1{,}170.0 & 27.4 & 28.0 & 18.7 & 15.99 & 0.558 \\
19 & LightGBM (tuned + ens.)  & 1{,}150.0 & 27.3 & 23.2 & 19.7 & 17.36 & 0.533 \\
22 & TabDPT-1.3               & 1{,}122.7 & 30.4 & 30.8 & 21.0 & 18.47 & 0.499 \\
25 & XGBoost (tuned + ens.)   & 1{,}112.9 & 27.5 & 24.4 & 21.5 & 19.00 & 0.487 \\
\bottomrule
\end{tabular}
\end{table}

\begin{figure}[H]
\centering
\includegraphics[width=0.85\textwidth]{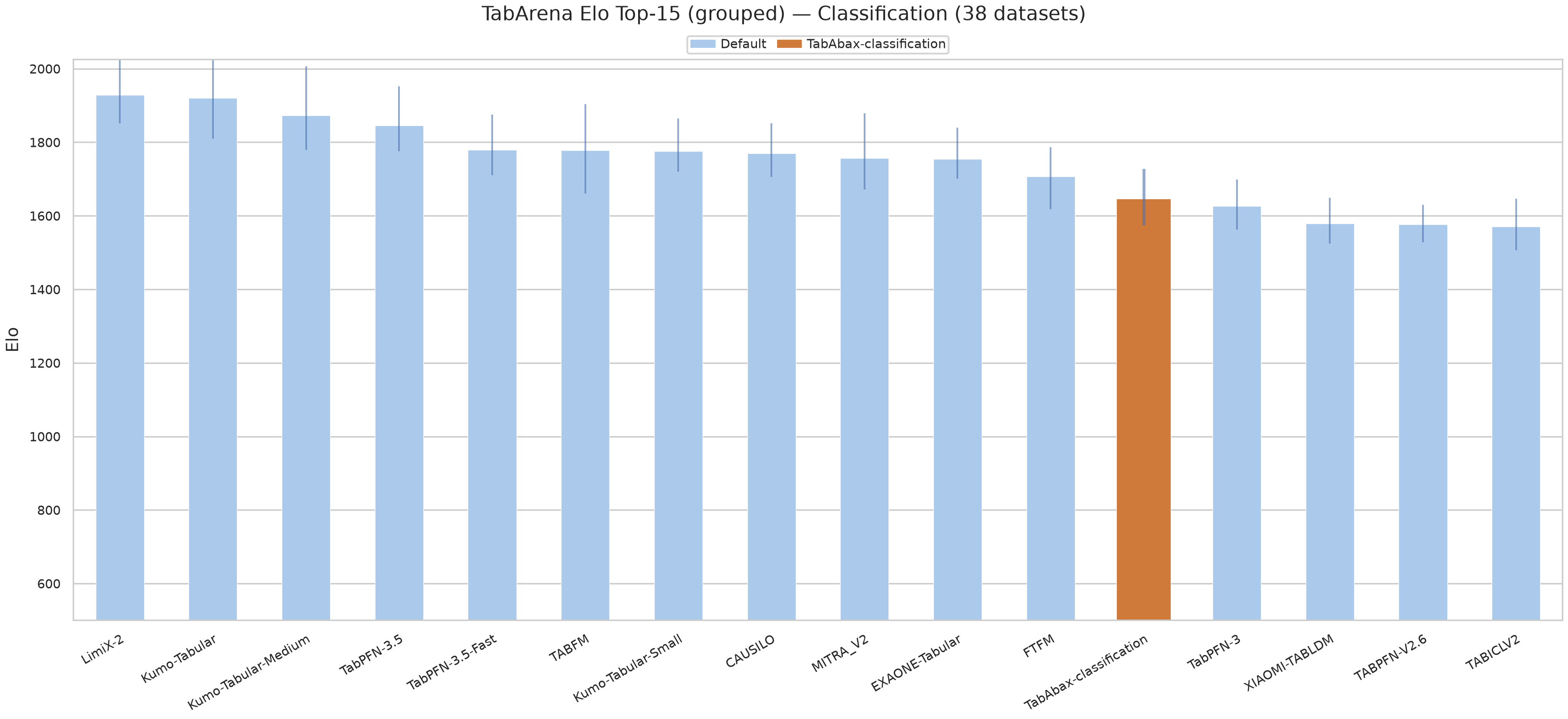}
\caption{TabArena classification Elo across all 38 classification datasets
(top-15 grouped leaderboard). \model{}-classification (dark orange bar)
reaches an Elo rating of 1{,}647, ranking 12th of all 99 evaluated methods, $+$76
over the TabICLv2 official default baseline (1{,}571) and $+$21 over
TabPFN-3 (1{,}626); the full 99-method table is in
Table~\ref{tab:app-classification}.}
\label{fig:tabarena-elo}
\end{figure}

\textbf{TabArena.} Table~\ref{tab:main} reports the overall leaderboard.
Figure~\ref{fig:tabarena-elo} reports the TabArena
classification Elo leaderboard; the complete 99-method tables for all three
suites are given in the appendix
(Tables~\ref{tab:app-overall},
\ref{tab:app-regression}, and~\ref{tab:app-classification}). Throughout,
TabArena entries appear under their submission names; our submission
appears as \texttt{TabAbax}, backed by the task-specific
checkpoints \texttt{TabAbax-classification} and
\texttt{TabAbax-regression} for the classification and
regression suites, respectively.

\begin{table}[H]
\centering
\caption{Overall TabArena leaderboard, selected methods (51 datasets; Elo
anchored at default Random Forest
$=1000$; Rank is among all 99 evaluated methods; metric definitions as
in Table~\ref{tab:beyond-main}). The
full 99-method table is in Table~\ref{tab:app-overall}.}
\label{tab:main}
\small
\begin{tabular}{c l r r r r r r}
\toprule
Rank & Method & Elo$\uparrow$ & Elo$+$ & Elo$-$ & Avg.\ rank$\downarrow$ & Improv.\ (\%)$\downarrow$ & Winrate$\uparrow$ \\
\midrule
6  & TabFM                    & 1{,}781.0 & 101.5 & 98.4 & 12.3 & 6.57  & 0.885 \\
7  & Causilo                  & 1{,}780.3 & 81.1  & 49.3 & 12.3 & 9.07  & 0.884 \\
11 & \textbf{\model{} (ours)} & \textbf{1{,}676.0} & \textbf{74.0} & \textbf{58.3} & \textbf{17.7} & \textbf{11.14} & \textbf{0.830} \\
14 & TabPFN-2.6               & 1{,}577.2 & 53.9  & 36.9 & 24.2 & 12.98 & 0.763 \\
16 & TabICLv2 (baseline)      & 1{,}561.0 & 55.8  & 52.0 & 25.4 & 12.47 & 0.751 \\
18 & TabDPT-1.3               & 1{,}517.2 & 57.4  & 41.1 & 28.9 & 14.21 & 0.715 \\
25 & LightGBM (tuned + ens.)  & 1{,}410.5 & 26.9  & 26.3 & 38.7 & 17.11 & 0.616 \\
36 & XGBoost (tuned + ens.)   & 1{,}358.9 & 28.7  & 29.0 & 43.9 & 17.82 & 0.562 \\
\bottomrule
\end{tabular}
\end{table}

\begin{figure}[H]
\centering
\includegraphics[width=0.85\textwidth]{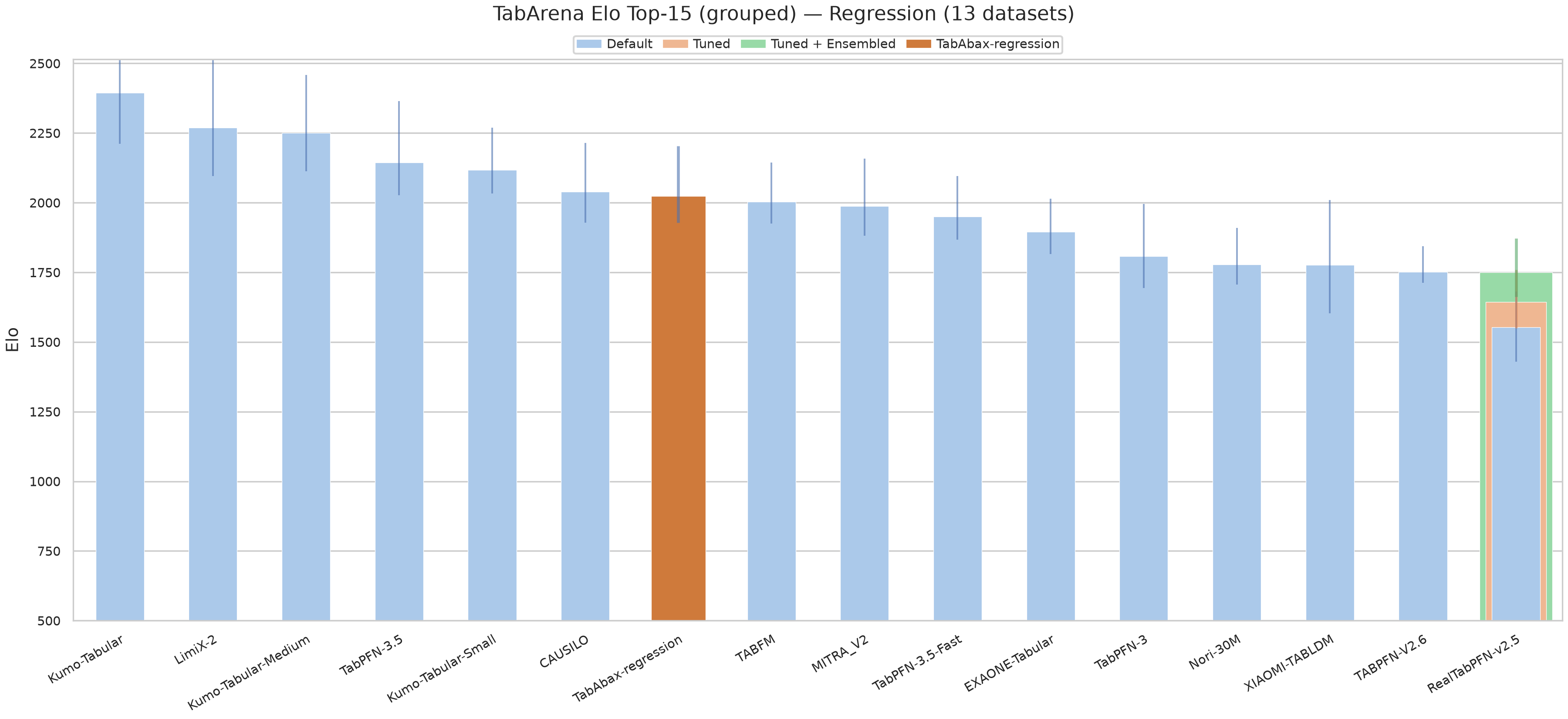}
\caption{TabArena regression Elo across all 13 datasets of the benchmark's
full regression suite (top-15 grouped leaderboard). \model{} (dark orange
bar; arena entry \texttt{TabAbax}, regression checkpoint) reaches an
Elo rating of 2{,}024, ranking 7th of all 99 evaluated methods, $+$336 over the
TabICLv2 official default baseline (1{,}688), $+$217 over TabPFN-3
(1{,}807), and $+$21 over TabFM (2{,}003). The full 99-method table is
in Table~\ref{tab:app-regression}.}
\label{fig:tabarena-reg-top15}
\end{figure}

\textbf{Regression.} \model{} performs particularly strongly on
regression: on all 13 regression datasets of TabArena---the benchmark's
full regression suite---our regression checkpoint reaches an Elo rating of
2{,}024, ranking 7th of all 99 evaluated methods
(Figure~\ref{fig:tabarena-reg-top15}; full table in
Table~\ref{tab:app-regression}), with an average per-dataset rank of 10.1,
a harmonic-mean rank of 6.3, a normalized score of 0.749, and a win rate
of 0.907. It surpasses
TabFM (2{,}003), Mitra-v2 (1{,}988), TabPFN-3.5-Fast (1{,}950), and
EXAONE-Tabular (1{,}896), and is behind only
Causilo (2{,}039) and the TabPFN-3.5 and Kumo family models---a $+$336 Elo
gain over the TabICLv2 baseline (1{,}688).
On the
oracle-improvability metric the position is identical: \model{} reaches
3.30\% and again ranks 7th,
on par with Mitra-v2 (3.33\%) and better than TabPFN-3.5-Fast (4.59\%),
TabPFN-3 (4.88\%), EXAONE-Tabular (5.15\%), and Xiaomi-TabLDM
(5.26\%)---as in classification, the rank is
stable across complementary metrics. The aggregate Score view agrees:
\model{} reaches 0.749, again 7th of the 99 methods.

\textbf{Classification.} On the 38 classification datasets of TabArena,
\model{} reaches an Elo rating of 1{,}647,
ranking 12th of all 99 evaluated methods
(Figure~\ref{fig:tabarena-elo}; full table in
Table~\ref{tab:app-classification})---a $+$76 Elo gain over the TabICLv2
baseline (1{,}571). The position is again stable across
complementary metrics: \model{} ranks 12th on \emph{all five} TabArena
metrics---Elo, oracle improvability (13.8\%), aggregate Score (0.555),
average rank (20.2), and harmonic rank (11.2)
(per-metric values in Table~\ref{tab:app-classification}).

\begin{figure}[H]
\centering
\includegraphics[width=0.72\textwidth]{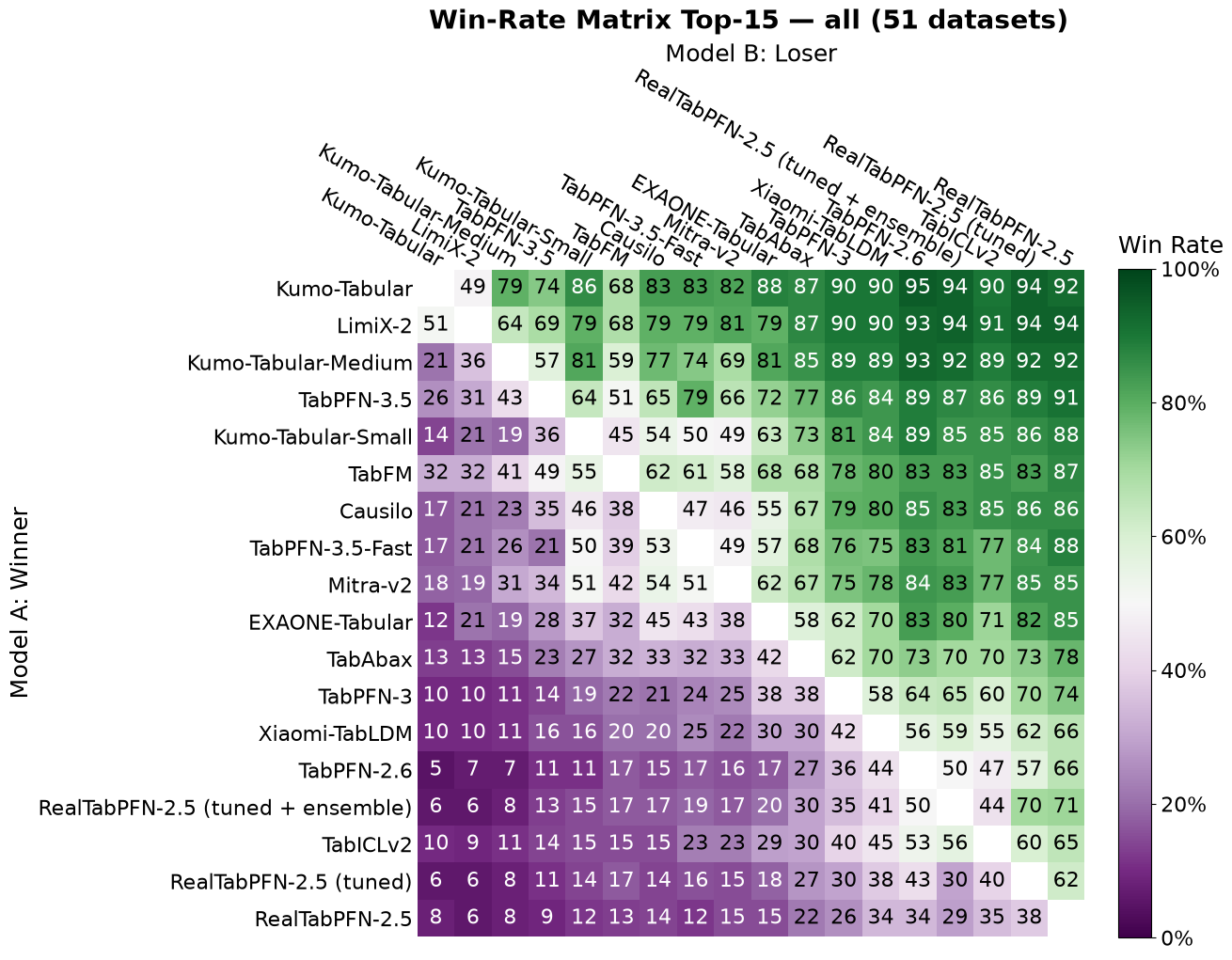}
\caption{Pairwise win-rate matrix among the top methods on the overall
TabArena suite (all 51 datasets). How to read it: locate \model{}'s
row and scan across the columns---each cell (row, column) is the
fraction of datasets on which the row method beats the column method
(darker green favors the row method, darker purple favors the column
method); reading down a column gives the mirror image, i.e., how often
that method loses to each row method. The
\model{} row shows wins against the TabICLv2 baseline on 78\% of
datasets, against TabPFN-3 on 70\%, and against Xiaomi-TabLDM,
TabPFN-v2.6, and the RealTabPFN-v2.5 variants on 62--73\%, while the
frontier group (Kumo family, TabPFN-3.5, TabFM, Causilo,
Mitra-v2, EXAONE-Tabular) remains ahead (13--42\%).}
\label{fig:winrate-matrix}
\end{figure}

\textbf{Overall performance.} Table~\ref{tab:main} reports the overall
leaderboard, and Figure~\ref{fig:tabarena-overall} shows the overall
TabArena Elo across all 51 datasets (the full 99-method table is
given in Table~\ref{tab:app-overall}):
\model{} reaches an Elo rating of
1{,}676, ranking 11th of the 99 evaluated methods, with a harmonic-mean
rank of 9.4, a median rank of 15, a normalized score of 0.60, and a win
rate of 0.83---$+$115 Elo above the TabICLv2
baseline (1{,}561). \model{} sits behind the
frontier group (Kumo family, TabPFN-3.5, TabFM, Causilo, Mitra-v2,
EXAONE-Tabular) and ahead of TabPFN-3 (1{,}626), Xiaomi-TabLDM
(1{,}583), TABPFN-v2.6 (1{,}577), and RealTabPFN-v2.5.
The pairwise win-rate matrix (Figure~\ref{fig:winrate-matrix}) shows
the same picture from the head-to-head view.

\subsection{Ablation: Isolating the Prior Changes}
\label{sec:ablation}
To disentangle the two axes of improvement, we additionally trained
prior-only variants of both checkpoints---the TabICLv2 architecture
left unchanged, with only the improved prior of Section~\ref{sec:prior}
enabled---and evaluated them under the same leaderboard protocol as
the full model (Table~\ref{tab:split}).

\begin{table}[H]
\centering
\caption{Classification vs.\ regression split under the official TabArena
metrics. The prior-only ablation rows (the unmodified TabICLv2
architecture trained with the improved prior;
Section~\ref{sec:ablation}) isolate the prior contribution to the
full model's $+$76 classification and $+$336 regression Elo gains over
the TabICLv2 baseline. Rank is the mean per-dataset rank, Improv.\ the
oracle improvability (\%, lower is better), and Win the per-dataset
win rate. Full per-suite tables: Tables~\ref{tab:app-classification}
and~\ref{tab:app-regression}.}
\label{tab:split}
\small
\setlength{\tabcolsep}{4pt}
\begin{tabular}{l rrrr rrrr}
\toprule
& \multicolumn{4}{c}{Classification (38 datasets)}
& \multicolumn{4}{c}{Regression (13 datasets)} \\
\cmidrule(lr){2-5}\cmidrule(lr){6-9}
Method & Elo$\uparrow$ & Rank$\downarrow$ & Improv.$\downarrow$ & Win$\uparrow$
& Elo$\uparrow$ & Rank$\downarrow$ & Improv.$\downarrow$ & Win$\uparrow$ \\
\midrule
\textbf{\model{} (ours)} & \textbf{1{,}646.7} & \textbf{20.2} & \textbf{13.83} & \textbf{0.804} & \textbf{2{,}023.7} & \textbf{10.1} & \textbf{3.30} & \textbf{0.907} \\
\model{}-prior\_only & 1{,}582.6 & 24.7 & 14.74 & 0.758 & 1{,}915.8 & 13.6 & 4.11 & 0.872 \\
TabICLv2 (baseline)     & 1{,}570.6 & 25.7 & 14.56 & 0.748 & 1{,}688.1 & 24.4 & 6.35 & 0.760 \\
\bottomrule
\end{tabular}
\end{table}

\begin{figure}[!htb]
\centering
\includegraphics[width=0.85\textwidth]{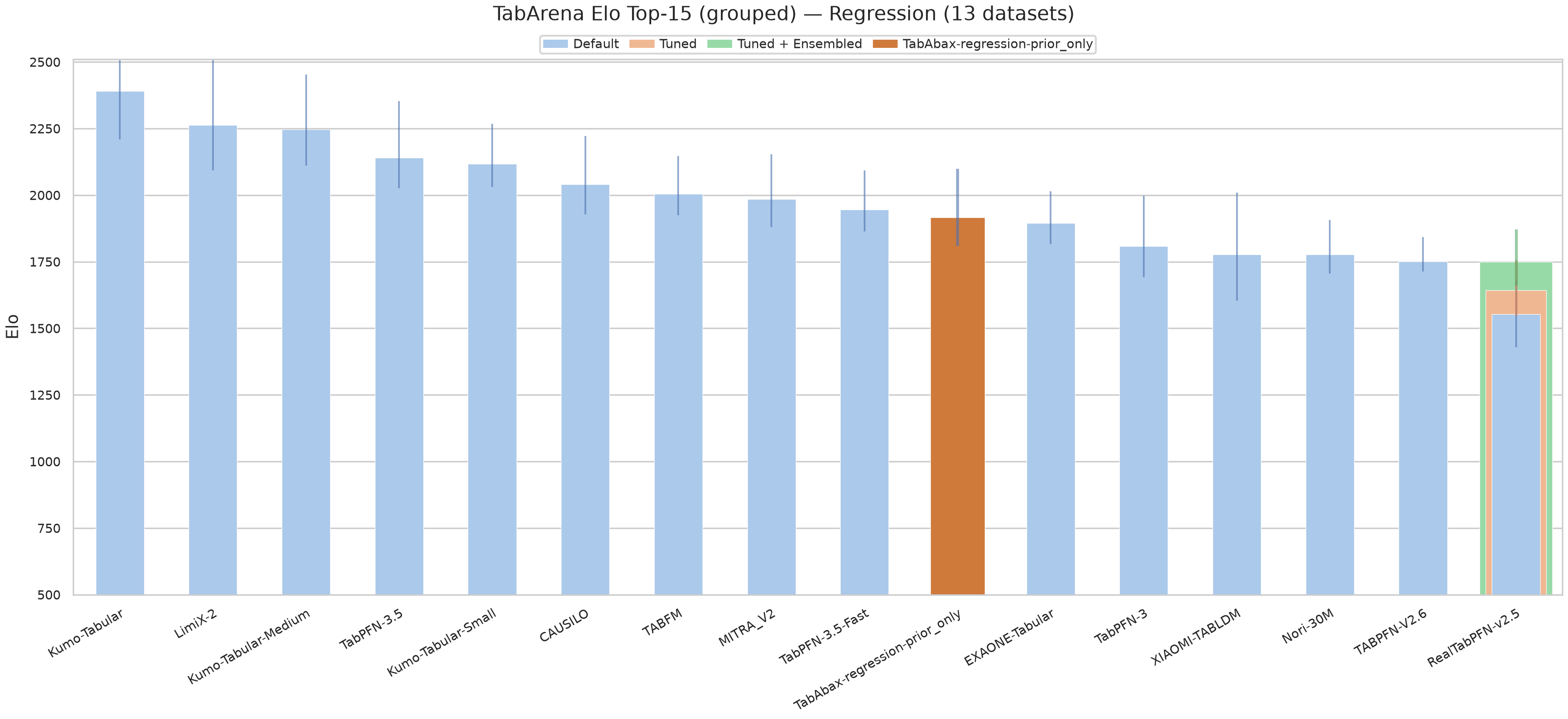}
\caption{Isolating the prior changes on the TabArena regression
suite (top-15 grouped leaderboard).
\texttt{TabAbax-regression-prior\_only} (dark orange
bar)---the unmodified TabICLv2 architecture trained with the improved prior---reaches an Elo rating of 1{,}916, $+$228 over the TabICLv2 official
default baseline (1{,}688), while the full \model{} regression
checkpoint (2{,}024; Figure~\ref{fig:tabarena-reg-top15}) adds $+$108
from the architectural changes.}
\label{fig:tabarena-reg-prioronly}
\end{figure}

\textbf{Regression.} \texttt{TabAbax-regression-prior\_only}
reaches an Elo rating of 1{,}916, ranking 10th of the
99 evaluated methods: $+$228 over the TabICLv2 official default
baseline (1{,}688), i.e., roughly two thirds of the full $+$336 gain,
while the remaining $+$108 come from the architectural changes
(Figure~\ref{fig:tabarena-reg-prioronly}). The prior-only variant thus
isolates the contribution of the prior changes and directly
supports the central claim that prior quality, not architectural
capacity, is the main lever for regression.

\begin{figure}[H]
\centering
\includegraphics[width=0.85\textwidth]{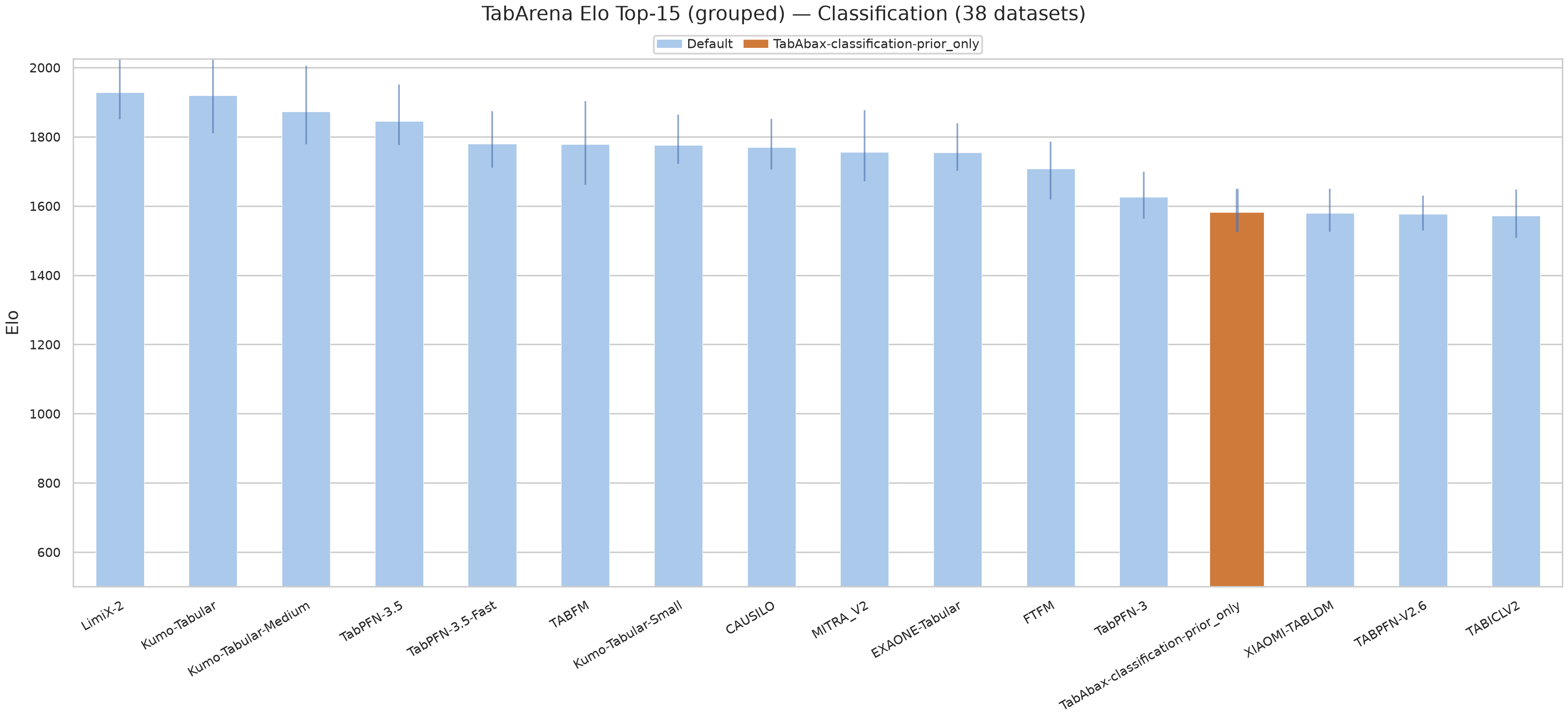}
\caption{Isolating the prior changes on the TabArena classification
suite (top-15 grouped leaderboard).
\texttt{TabAbax-classification-prior\_only} (dark orange
bar)---the unmodified TabICLv2 architecture trained with the improved prior---reaches an Elo rating of 1{,}583, $+$12 over the TabICLv2 official
default baseline (1{,}571), while the full \model{} classification
checkpoint (1{,}647; Figure~\ref{fig:tabarena-elo}) adds $+$64 from
the architectural changes.}
\label{fig:tabarena-clf-prioronly}
\end{figure}

{\sloppypar
\textbf{Classification.} The same decomposition mirrors the regression
result: \texttt{TabAbax-classification-prior\_only} reaches an
Elo rating of 1{,}583, ranking 13th of the 99 evaluated methods, only $+$12
over the TabICLv2 baseline
(1{,}571), while the full \model{} classification checkpoint (1{,}647;
Figure~\ref{fig:tabarena-elo}) adds $+$64 from the architectural
changes (Figure~\ref{fig:tabarena-clf-prioronly}). The two suites are
thus mirror images: the regression gain is dominated by the improved
prior, and the classification gain by the architectural
changes. The small but nonzero prior-side gain on classification
($+$12) is consistent with what the prior changes act on: strict
one-hot width matching and Dirichlet-skewed class frequencies shape
the categorical representations and class priors that classification
consumes directly, whereas the native smoothness of continuous
targets---the aspect most relevant to regression---does not affect
classification decisions.\par}

\section{Limitations}
\label{sec:limitations}
\textbf{Limitations of the current evaluation.}
The predictability
filters reduce the effective corpus size, and we have not yet
measured whether the diversity lost by filtering trades off against the
quality gained at the extremes of the distribution. Beyond corpus size,
filtering risks a coverage gap for \emph{weak-signal, high-noise}
tasks---exactly the regime of many real-world datasets, which resemble
the problems the filter removes---so generalization on weak-signal
real data may be reduced.

\noindent\textbf{Context and memory bounds.} The ICL context holds a bounded
number of rows; very large datasets and long-range dependencies remain
limited by context size and memory. And like all in-context learners,
predictions depend on the quality of the context rows: corrupted,
mislabeled, or distribution-shifted demonstrations can silently
degrade predictions.

\section{Conclusion}
\model{} extends TabICLv2 along two upstream axes. On the prior side,
generation-time changes concentrate on data quality and complexity:
strict one-hot width matching, class frequencies drawn from multiple
distributions, richer converter function families, and the removal of
the post-hoc monotone warping that artificially introduces
non-smoothness into regression targets. On the encoder side, Fourier
scalar encoding and a two-pass
column--row stack adapt the pre-compression recipe of TabFM and Causilo
to TabICLv2's own parameter budget. The prior-side changes yield large
regression gains without increasing model parameter count or altering
the inference architecture, the architectural changes
benefit both task types, and the combination carries both gains at
matched asymptotic inference cost. We believe
prior quality control is a generally applicable recipe for
SCM-pretrained
tabular foundation models. More broadly, our results suggest that for
such models, regression performance can be
substantially improved through prior quality control, without
scaling up model size.

\section{Future Work}
\textbf{Deeper post-compression interaction.} The encoder evolution
from TabICL's single pre-compression pass, to TabFM's two untied
passes, to Kumo's repeated post-compression column--row interaction
suggests that interaction \emph{after} compression is the next
promising axis; we plan to move in this direction, adding deeper
pre- and post-compression interaction stages within the linear-time
inference budget.

\noindent\textbf{Width and capacity scaling.} The changes above are all
established at TabICLv2's original width and parameter budget. We
plan to scale the model---enlarging the hidden dimension of the
column--row encoder and ICL backbone together with the parameter
count---and study whether the gains from the prior changes and
deeper pre-compression interaction persist, amplify, or saturate at
larger capacity.

\subsection*{Model parameter counts}
As shown in Table~\ref{tab:params}, we list the parameter counts and
checkpoint sizes of the
released \model{} checkpoints. \model{} is a dense architecture (no sparse
mixture-of-experts routing), so all parameters are active in a single
forward pass; the classifier and regressor share the same encoder stack
and differ only in the output head (class logits vs.\ 999 predicted
quantiles).

\begin{table}[H]
\centering
\caption{Parameter counts and checkpoint sizes of the released
\model{} checkpoints.}
\label{tab:params}
\small
\renewcommand{\arraystretch}{1.15}
\begin{tabular*}{\textwidth}{@{\extracolsep{\fill}} l l r r}
\toprule
Model & Type & Total params & File size \\
\midrule
\texttt{TabAbax-classifier.ckpt} & Classification (\texttt{max\_classes}=10) & 28.85M & 231.15\,MB \\
\texttt{TabAbax-regressor.ckpt}  & Regression (\texttt{quantiles}=999)        & 29.84M & 239.03\,MB \\
\bottomrule
\end{tabular*}
\end{table}

\clearpage
\appendix
\section{Full leaderboard tables}
\label{app:leaderboards}

Tables~\ref{tab:app-beyondarena}, \ref{tab:app-beyondarena-reg},
and~\ref{tab:app-beyondarena-cls} report the complete BeyondArena
leaderboards underlying Figures~\ref{fig:beyondarena-all-top15},
\ref{fig:beyondarena-regression},
and~\ref{fig:beyondarena-classification};
Tables~\ref{tab:app-overall}, \ref{tab:app-regression}, and
~\ref{tab:app-classification} report the complete TabArena
leaderboards underlying Figures~\ref{fig:tabarena-reg-top15},
\ref{fig:tabarena-overall}, and~\ref{fig:tabarena-elo}
(leaderboard snapshot as of October 2026).
\model{} is evaluated as a single submission (\texttt{TabAbax});
Elo$+$ / Elo$-$ are the upper/lower error-bar extents, Avg.\ rank and
Harm.\ rank are the mean and harmonic-mean per-dataset ranks, Score is the
normalized aggregate score, Improv.\ is the oracle improvability (gap to the
per-dataset best method, lower is better), and Winrate is the per-dataset
win rate against all other methods. \textbf{Bold}: our submission.

{\scriptsize
\setlength{\tabcolsep}{3.2pt}
\begin{longtable}{l r r r r r r r r}
\caption{Full BeyondArena leaderboard: all 142 datasets (98 classification and 44 regression), 41 evaluated methods, sorted by Elo.}\label{tab:app-beyondarena}\\
\toprule
Method & Elo & Elo$+$ & Elo$-$ & Avg.\ rank & Harm.\ rank & Score & Improv.\ (\%) & Winrate \\
\midrule
\endfirsthead
\multicolumn{9}{l}{\footnotesize\emph{Table \ref{tab:app-beyondarena} (continued)}}\\
\toprule
Method & Elo & Elo$+$ & Elo$-$ & Avg.\ rank & Harm.\ rank & Score & Improv.\ (\%) & Winrate \\
\midrule
\endhead
\midrule
\multicolumn{9}{r}{\footnotesize continued on next page}\\
\midrule
\endfoot
\bottomrule
\endlastfoot
Kumo-Tabular (default) & 1{,}429.4 & 42.5 & 43.4 & 7.8 & 2.7 & 0.747 & 7.77 & 0.830 \\
Kumo-Tabular-Medium (default) & 1{,}378.4 & 40.0 & 39.7 & 9.5 & 4.3 & 0.669 & 9.26 & 0.788 \\
TabPFN-3.5 (default) & 1{,}377.1 & 39.6 & 35.5 & 9.5 & 4.2 & 0.654 & 11.04 & 0.786 \\
Causilo (default) & 1{,}328.7 & 30.7 & 26.8 & 11.4 & 6.6 & 0.558 & 12.03 & 0.740 \\
Kumo-Tabular-Small (default) & 1{,}323.0 & 28.6 & 30.2 & 11.6 & 6.8 & 0.546 & 11.81 & 0.734 \\
TabPFN-3.5-Fast (default) & 1{,}320.6 & 34.8 & 25.3 & 11.7 & 6.1 & 0.560 & 12.14 & 0.732 \\
LimiX-2 (default) & 1{,}288.7 & 51.7 & 45.4 & 13.1 & 3.1 & 0.570 & 10.09 & 0.698 \\
TabFM (default) & 1{,}274.4 & 39.4 & 36.9 & 13.7 & 4.6 & 0.497 & 10.82 & 0.682 \\
EXAONE-Tabular (default) & 1{,}245.6 & 31.7 & 33.8 & 15.0 & 7.8 & 0.464 & 14.19 & 0.649 \\
\textbf{TabAbax (default)} & \textbf{1{,}237.7} & \textbf{29.0} & \textbf{29.1} & \textbf{15.4} & \textbf{8.8} & \textbf{0.411} & \textbf{14.47} & \textbf{0.640} \\
TabPFN-3 (default) & 1{,}232.6 & 30.2 & 26.0 & 15.6 & 8.7 & 0.382 & 14.55 & 0.634 \\
RealMLP (tuned + ensembled) & 1{,}225.3 & 26.9 & 21.1 & 16.0 & 8.4 & 0.305 & 15.49 & 0.625 \\
TabPFN-2.6 (default) & 1{,}175.9 & 26.4 & 24.0 & 18.4 & 11.8 & 0.292 & 16.56 & 0.565 \\
TabM (tuned + ensembled) & 1{,}172.1 & 26.8 & 25.9 & 18.6 & 11.3 & 0.215 & 16.78 & 0.561 \\
CatBoost (tuned + ensembled) & 1{,}171.8 & 28.7 & 21.0 & 18.6 & 10.8 & 0.187 & 17.02 & 0.560 \\
TabICLv2 (default) & 1{,}170.0 & 27.4 & 28.0 & 18.7 & 10.6 & 0.318 & 15.99 & 0.558 \\
RealMLP (tuned) & 1{,}167.0 & 27.8 & 23.5 & 18.8 & 9.5 & 0.227 & 16.75 & 0.554 \\
CatBoost (tuned) & 1{,}158.3 & 28.3 & 21.1 & 19.3 & 11.6 & 0.178 & 17.19 & 0.544 \\
LightGBM (tuned + ensembled) & 1{,}150.0 & 27.3 & 23.2 & 19.7 & 10.3 & 0.182 & 17.36 & 0.533 \\
RealTabPFN-2.5 (default) & 1{,}141.5 & 27.3 & 26.9 & 20.1 & 13.0 & 0.240 & 17.56 & 0.522 \\
TabM (tuned) & 1{,}138.8 & 22.3 & 25.2 & 20.2 & 12.5 & 0.166 & 17.51 & 0.519 \\
TabDPT-1.3 (default) & 1{,}122.7 & 30.4 & 30.8 & 21.0 & 11.5 & 0.257 & 18.47 & 0.499 \\
LightGBM (tuned) & 1{,}121.3 & 27.5 & 25.4 & 21.1 & 11.6 & 0.157 & 17.89 & 0.497 \\
CatBoost (default) & 1{,}117.4 & 28.6 & 24.4 & 21.3 & 13.7 & 0.151 & 18.34 & 0.492 \\
XGBoost (tuned + ensembled) & 1{,}112.9 & 27.5 & 24.4 & 21.5 & 13.1 & 0.156 & 19.00 & 0.487 \\
XGBoost (tuned) & 1{,}092.0 & 26.8 & 25.3 & 22.6 & 14.7 & 0.129 & 19.30 & 0.461 \\
TabM (default) & 1{,}091.2 & 25.7 & 29.2 & 22.6 & 14.9 & 0.127 & 18.93 & 0.460 \\
RealMLP (default) & 1{,}047.7 & 27.8 & 32.4 & 24.8 & 14.7 & 0.109 & 20.26 & 0.406 \\
TabSwift (default) & 1{,}047.4 & 28.5 & 30.1 & 24.8 & 15.3 & 0.124 & 21.18 & 0.406 \\
TabDPT (default) & 1{,}027.7 & 26.7 & 36.1 & 25.7 & 15.0 & 0.149 & 22.35 & 0.382 \\
XGBoost (default) & 1{,}000.0 & 29.8 & 28.5 & 27.0 & 20.4 & 0.081 & 21.84 & 0.350 \\
LightGBM (default) & 989.8 & 28.0 & 23.4 & 27.5 & 19.8 & 0.063 & 21.88 & 0.338 \\
ExtraTrees (tuned + ensembled) & 989.6 & 23.1 & 25.6 & 27.5 & 20.3 & 0.033 & 22.45 & 0.338 \\
ExtraTrees (tuned) & 956.4 & 26.7 & 29.2 & 29.0 & 21.0 & 0.029 & 23.18 & 0.301 \\
RandomForest (tuned + ensembled) & 951.0 & 25.3 & 32.2 & 29.2 & 23.2 & 0.015 & 23.49 & 0.295 \\
RandomForest (tuned) & 933.7 & 24.7 & 31.9 & 29.9 & 23.2 & 0.013 & 23.85 & 0.277 \\
Linear (tuned + ensembled) & 908.3 & 36.2 & 41.6 & 31.0 & 18.5 & 0.064 & 29.01 & 0.251 \\
Linear (tuned) & 873.7 & 38.7 & 46.1 & 32.3 & 19.1 & 0.048 & 30.06 & 0.218 \\
RandomForest (default) & 873.3 & 31.0 & 32.2 & 32.3 & 25.5 & 0.012 & 26.14 & 0.218 \\
ExtraTrees (default) & 861.5 & 38.6 & 38.5 & 32.7 & 23.9 & 0.030 & 26.89 & 0.208 \\
Linear (default) & 804.6 & 43.0 & 51.0 & 34.5 & 22.3 & 0.037 & 33.01 & 0.161 \\
\end{longtable}
}

{\scriptsize
\setlength{\tabcolsep}{3.2pt}
\begin{longtable}{l r r r r r r r r}
\caption{Full BeyondArena regression leaderboard: all 44 regression datasets, 41 evaluated methods, sorted by Elo.}\label{tab:app-beyondarena-reg}\\
\toprule
Method & Elo & Elo$+$ & Elo$-$ & Avg.\ rank & Harm.\ rank & Score & Improv.\ (\%) & Winrate \\
\midrule
\endfirsthead
\multicolumn{9}{l}{\footnotesize\emph{Table \ref{tab:app-beyondarena-reg} (continued)}}\\
\toprule
Method & Elo & Elo$+$ & Elo$-$ & Avg.\ rank & Harm.\ rank & Score & Improv.\ (\%) & Winrate \\
\midrule
\endhead
\midrule
\multicolumn{9}{r}{\footnotesize continued on next page}\\
\midrule
\endfoot
\bottomrule
\endlastfoot
Kumo-Tabular (default) & 1{,}497.5 & 111.0 & 80.3 & 6.2 & 2.3 & 0.802 & 3.52 & 0.870 \\
Kumo-Tabular-Medium (default) & 1{,}436.9 & 89.6 & 58.1 & 7.9 & 3.9 & 0.721 & 3.81 & 0.828 \\
TabPFN-3.5 (default) & 1{,}383.0 & 75.1 & 61.0 & 9.7 & 4.3 & 0.611 & 5.92 & 0.783 \\
Kumo-Tabular-Small (default) & 1{,}376.1 & 80.0 & 47.6 & 9.9 & 5.9 & 0.589 & 5.14 & 0.777 \\
Causilo (default) & 1{,}341.2 & 82.8 & 56.9 & 11.2 & 6.2 & 0.529 & 6.33 & 0.744 \\
LimiX-2 (default) & 1{,}293.5 & 109.1 & 91.4 & 13.2 & 2.7 & 0.583 & 6.08 & 0.695 \\
TabPFN-3.5-Fast (default) & 1{,}288.9 & 68.3 & 57.9 & 13.4 & 6.9 & 0.448 & 7.32 & 0.690 \\
\textbf{TabAbax (default)} & \textbf{1{,}286.7} & \textbf{64.7} & \textbf{43.2} & \textbf{13.5} & \textbf{7.9} & \textbf{0.450} & \textbf{7.29} & \textbf{0.688} \\
RealMLP (tuned + ensembled) & 1{,}274.0 & 53.1 & 37.9 & 14.0 & 6.4 & 0.374 & 6.21 & 0.674 \\
EXAONE-Tabular (default) & 1{,}248.0 & 66.2 & 68.8 & 15.2 & 7.7 & 0.456 & 8.23 & 0.645 \\
TabPFN-3 (default) & 1{,}236.4 & 40.2 & 42.6 & 15.7 & 10.5 & 0.334 & 7.55 & 0.632 \\
TabFM (default) & 1{,}233.7 & 82.5 & 67.0 & 15.8 & 8.7 & 0.375 & 8.01 & 0.629 \\
RealMLP (tuned) & 1{,}218.6 & 51.7 & 39.9 & 16.5 & 7.3 & 0.300 & 6.78 & 0.612 \\
TabPFN-2.6 (default) & 1{,}182.5 & 66.2 & 53.3 & 18.2 & 11.8 & 0.308 & 9.20 & 0.569 \\
CatBoost (tuned + ensembled) & 1{,}181.8 & 45.9 & 41.0 & 18.3 & 10.1 & 0.208 & 9.09 & 0.568 \\
TabM (tuned + ensembled) & 1{,}177.8 & 41.6 & 47.5 & 18.5 & 11.4 & 0.224 & 8.12 & 0.564 \\
LightGBM (tuned + ensembled) & 1{,}170.0 & 48.0 & 47.1 & 18.8 & 9.6 & 0.227 & 7.86 & 0.554 \\
CatBoost (tuned) & 1{,}161.8 & 42.0 & 39.5 & 19.2 & 12.5 & 0.192 & 9.26 & 0.544 \\
TabICLv2 (default) & 1{,}159.5 & 62.3 & 58.0 & 19.3 & 12.2 & 0.261 & 9.75 & 0.542 \\
LightGBM (tuned) & 1{,}144.0 & 50.0 & 49.1 & 20.1 & 11.3 & 0.199 & 8.17 & 0.523 \\
TabM (tuned) & 1{,}140.1 & 39.0 & 45.9 & 20.3 & 13.5 & 0.154 & 8.96 & 0.518 \\
RealTabPFN-2.5 (default) & 1{,}128.2 & 57.0 & 60.7 & 20.8 & 14.2 & 0.220 & 10.00 & 0.504 \\
XGBoost (tuned + ensembled) & 1{,}124.9 & 44.5 & 40.8 & 21.0 & 14.8 & 0.179 & 10.25 & 0.500 \\
CatBoost (default) & 1{,}116.1 & 45.2 & 52.1 & 21.4 & 14.0 & 0.179 & 10.44 & 0.489 \\
XGBoost (tuned) & 1{,}099.5 & 37.9 & 44.8 & 22.2 & 15.6 & 0.144 & 10.45 & 0.470 \\
RealMLP (default) & 1{,}099.0 & 46.8 & 48.9 & 22.2 & 14.5 & 0.153 & 9.51 & 0.469 \\
TabDPT-1.3 (default) & 1{,}091.3 & 47.2 & 55.9 & 22.6 & 12.9 & 0.229 & 11.97 & 0.460 \\
TabM (default) & 1{,}083.0 & 40.1 & 55.3 & 23.0 & 15.3 & 0.101 & 11.24 & 0.450 \\
TabDPT (default) & 1{,}069.1 & 51.2 & 59.7 & 23.7 & 13.5 & 0.201 & 12.77 & 0.434 \\
TabSwift (default) & 1{,}025.4 & 50.6 & 54.7 & 25.7 & 16.2 & 0.124 & 12.98 & 0.383 \\
LightGBM (default) & 1{,}011.8 & 42.1 & 52.8 & 26.3 & 21.3 & 0.090 & 11.70 & 0.368 \\
XGBoost (default) & 1{,}000.0 & 46.1 & 58.4 & 26.8 & 21.4 & 0.104 & 12.33 & 0.355 \\
ExtraTrees (tuned + ensembled) & 967.4 & 49.7 & 53.3 & 28.2 & 22.6 & 0.028 & 13.46 & 0.320 \\
ExtraTrees (tuned) & 947.8 & 52.8 & 52.4 & 29.0 & 22.1 & 0.023 & 13.58 & 0.300 \\
RandomForest (tuned + ensembled) & 923.5 & 53.4 & 54.5 & 30.0 & 26.2 & 0.009 & 14.36 & 0.276 \\
RandomForest (tuned) & 916.9 & 48.4 & 56.8 & 30.2 & 26.2 & 0.006 & 14.39 & 0.269 \\
ExtraTrees (default) & 869.7 & 66.1 & 72.1 & 31.9 & 25.0 & 0.011 & 14.94 & 0.227 \\
RandomForest (default) & 829.5 & 45.6 & 70.9 & 33.3 & 27.3 & 0.002 & 16.10 & 0.193 \\
Linear (tuned + ensembled) & 793.4 & 77.1 & 92.9 & 34.3 & 25.0 & 0.010 & 24.48 & 0.167 \\
Linear (tuned) & 741.3 & 85.0 & 103.2 & 35.7 & 26.5 & 0.008 & 25.25 & 0.132 \\
Linear (default) & 647.0 & 95.6 & 116.4 & 37.7 & 26.4 & 0.005 & 28.49 & 0.083 \\
\end{longtable}
}

{\scriptsize
\setlength{\tabcolsep}{3.2pt}
\begin{longtable}{l r r r r r r r r}
\caption{Full BeyondArena classification leaderboard: all 98 classification datasets, 41 evaluated methods, sorted by Elo.}\label{tab:app-beyondarena-cls}\\
\toprule
Method & Elo & Elo$+$ & Elo$-$ & Avg.\ rank & Harm.\ rank & Score & Improv.\ (\%) & Winrate \\
\midrule
\endfirsthead
\multicolumn{9}{l}{\footnotesize\emph{Table \ref{tab:app-beyondarena-cls} (continued)}}\\
\toprule
Method & Elo & Elo$+$ & Elo$-$ & Avg.\ rank & Harm.\ rank & Score & Improv.\ (\%) & Winrate \\
\midrule
\endhead
\midrule
\multicolumn{9}{r}{\footnotesize continued on next page}\\
\midrule
\endfoot
\bottomrule
\endlastfoot
Kumo-Tabular (default) & 1{,}405.1 & 60.0 & 54.3 & 8.5 & 2.9 & 0.722 & 9.68 & 0.813 \\
TabPFN-3.5 (default) & 1{,}376.8 & 44.5 & 39.9 & 9.5 & 4.1 & 0.673 & 13.34 & 0.788 \\
Kumo-Tabular-Medium (default) & 1{,}357.0 & 54.6 & 40.2 & 10.2 & 4.5 & 0.645 & 11.70 & 0.770 \\
TabPFN-3.5-Fast (default) & 1{,}337.3 & 37.2 & 35.6 & 11.0 & 5.8 & 0.610 & 14.30 & 0.750 \\
Causilo (default) & 1{,}325.2 & 30.2 & 33.0 & 11.5 & 6.8 & 0.571 & 14.59 & 0.738 \\
Kumo-Tabular-Small (default) & 1{,}303.2 & 35.6 & 35.6 & 12.4 & 7.3 & 0.526 & 14.81 & 0.715 \\
TabFM (default) & 1{,}294.7 & 54.9 & 44.3 & 12.8 & 3.8 & 0.552 & 12.08 & 0.706 \\
LimiX-2 (default) & 1{,}288.3 & 60.1 & 58.9 & 13.0 & 3.4 & 0.564 & 11.89 & 0.699 \\
EXAONE-Tabular (default) & 1{,}246.1 & 40.7 & 37.0 & 15.0 & 7.8 & 0.468 & 16.87 & 0.651 \\
TabPFN-3 (default) & 1{,}232.5 & 27.9 & 33.6 & 15.6 & 8.1 & 0.404 & 17.70 & 0.635 \\
\textbf{TabAbax (default)} & \textbf{1{,}218.8} & \textbf{30.2} & \textbf{33.7} & \textbf{16.3} & \textbf{9.4} & \textbf{0.393} & \textbf{17.70} & \textbf{0.618} \\
RealMLP (tuned + ensembled) & 1{,}206.4 & 23.9 & 26.5 & 16.9 & 9.8 & 0.274 & 19.66 & 0.603 \\
TabICLv2 (default) & 1{,}175.6 & 31.3 & 40.7 & 18.4 & 10.0 & 0.344 & 18.79 & 0.565 \\
TabPFN-2.6 (default) & 1{,}174.2 & 28.2 & 29.7 & 18.5 & 11.8 & 0.284 & 19.87 & 0.564 \\
TabM (tuned + ensembled) & 1{,}170.8 & 24.4 & 31.5 & 18.6 & 11.3 & 0.211 & 20.67 & 0.559 \\
CatBoost (tuned + ensembled) & 1{,}168.7 & 23.5 & 25.3 & 18.7 & 11.1 & 0.178 & 20.59 & 0.557 \\
CatBoost (tuned) & 1{,}157.9 & 24.8 & 26.3 & 19.3 & 11.3 & 0.172 & 20.75 & 0.543 \\
RealTabPFN-2.5 (default) & 1{,}148.1 & 37.8 & 21.9 & 19.8 & 12.5 & 0.249 & 20.95 & 0.531 \\
RealMLP (tuned) & 1{,}146.4 & 28.8 & 27.7 & 19.9 & 11.1 & 0.194 & 21.23 & 0.529 \\
LightGBM (tuned + ensembled) & 1{,}142.6 & 28.4 & 25.9 & 20.1 & 10.7 & 0.162 & 21.62 & 0.524 \\
TabM (tuned) & 1{,}139.3 & 23.6 & 31.0 & 20.2 & 12.0 & 0.171 & 21.34 & 0.520 \\
TabDPT-1.3 (default) & 1{,}136.8 & 37.2 & 35.8 & 20.3 & 11.0 & 0.269 & 21.40 & 0.516 \\
CatBoost (default) & 1{,}118.8 & 28.2 & 28.7 & 21.3 & 13.5 & 0.139 & 21.89 & 0.494 \\
LightGBM (tuned) & 1{,}112.5 & 31.1 & 26.7 & 21.6 & 11.7 & 0.138 & 22.26 & 0.486 \\
XGBoost (tuned + ensembled) & 1{,}108.7 & 30.5 & 33.5 & 21.8 & 12.5 & 0.145 & 22.93 & 0.481 \\
TabM (default) & 1{,}095.3 & 26.4 & 30.5 & 22.4 & 14.7 & 0.139 & 22.38 & 0.464 \\
XGBoost (tuned) & 1{,}089.5 & 31.2 & 33.2 & 22.7 & 14.3 & 0.123 & 23.27 & 0.457 \\
TabSwift (default) & 1{,}056.9 & 28.1 & 29.2 & 24.4 & 14.9 & 0.123 & 24.86 & 0.416 \\
RealMLP (default) & 1{,}025.8 & 36.9 & 34.7 & 25.9 & 14.8 & 0.089 & 25.08 & 0.378 \\
TabDPT (default) & 1{,}010.0 & 37.0 & 32.2 & 26.6 & 15.8 & 0.126 & 26.65 & 0.359 \\
XGBoost (default) & 1{,}000.0 & 29.6 & 37.1 & 27.1 & 19.9 & 0.071 & 26.11 & 0.347 \\
ExtraTrees (tuned + ensembled) & 998.6 & 27.8 & 29.1 & 27.2 & 19.4 & 0.035 & 26.48 & 0.346 \\
LightGBM (default) & 980.3 & 33.9 & 35.6 & 28.0 & 19.1 & 0.052 & 26.46 & 0.324 \\
RandomForest (tuned + ensembled) & 961.7 & 31.2 & 31.9 & 28.9 & 22.0 & 0.017 & 27.58 & 0.304 \\
ExtraTrees (tuned) & 959.6 & 32.4 & 33.7 & 28.9 & 20.5 & 0.031 & 27.49 & 0.301 \\
Linear (tuned + ensembled) & 948.5 & 40.1 & 45.8 & 29.4 & 16.6 & 0.088 & 31.04 & 0.289 \\
RandomForest (tuned) & 939.9 & 30.0 & 29.2 & 29.8 & 22.0 & 0.016 & 28.10 & 0.280 \\
Linear (tuned) & 917.8 & 44.2 & 50.6 & 30.7 & 17.0 & 0.066 & 32.22 & 0.257 \\
RandomForest (default) & 889.3 & 35.0 & 44.8 & 31.8 & 24.8 & 0.017 & 30.66 & 0.229 \\
ExtraTrees (default) & 856.1 & 41.2 & 57.8 & 33.0 & 23.4 & 0.039 & 32.25 & 0.199 \\
Linear (default) & 853.0 & 51.4 & 58.1 & 33.1 & 20.8 & 0.051 & 35.04 & 0.196 \\
\end{longtable}
}

{\scriptsize
\setlength{\tabcolsep}{3.2pt}
\begin{longtable}{l r r r r r r r r}
\caption{Full TabArena overall leaderboard: all 51 datasets (38 classification and 13 regression), 99 evaluated methods, sorted by Elo.}\label{tab:app-overall}\\
\toprule
Method & Elo & Elo$+$ & Elo$-$ & Avg.\ rank & Harm.\ rank & Score & Improv.\ (\%) & Winrate \\
\midrule
\endfirsthead
\multicolumn{9}{l}{\footnotesize\emph{Table \ref{tab:app-overall} (continued)}}\\
\toprule
Method & Elo & Elo$+$ & Elo$-$ & Avg.\ rank & Harm.\ rank & Score & Improv.\ (\%) & Winrate \\
\midrule
\endhead
\midrule
\multicolumn{9}{r}{\footnotesize continued on next page}\\
\midrule
\endfoot
\bottomrule
\endlastfoot
Kumo-Tabular (default) & 1{,}957.2 & 160.7 & 99.5 & 6.2 & 2.4 & 0.885 & 4.39 & 0.946 \\
LimiX-2 (default) & 1{,}948.0 & 128.3 & 82.8 & 6.5 & 2.2 & 0.903 & 3.32 & 0.944 \\
Kumo-Tabular-Medium (default) & 1{,}898.1 & 105.0 & 77.4 & 7.9 & 4.2 & 0.827 & 5.88 & 0.929 \\
TabPFN-3.5 (default) & 1{,}859.1 & 102.0 & 65.2 & 9.2 & 4.2 & 0.784 & 7.56 & 0.916 \\
Kumo-Tabular-Small (default) & 1{,}797.4 & 79.5 & 56.1 & 11.6 & 7.5 & 0.721 & 8.90 & 0.892 \\
TabFM (default) & 1{,}781.0 & 101.5 & 98.4 & 12.3 & 3.7 & 0.738 & 6.57 & 0.885 \\
Causilo (default) & 1{,}780.3 & 81.1 & 49.3 & 12.3 & 7.0 & 0.708 & 9.07 & 0.884 \\
TabPFN-3.5-Fast (default) & 1{,}772.0 & 84.1 & 55.3 & 12.7 & 7.0 & 0.701 & 8.95 & 0.881 \\
Mitra-v2 (default) & 1{,}760.9 & 92.2 & 67.7 & 13.2 & 6.0 & 0.701 & 8.37 & 0.875 \\
EXAONE-Tabular (default) & 1{,}741.8 & 68.3 & 49.9 & 14.1 & 7.7 & 0.675 & 9.65 & 0.866 \\
\textbf{TabAbax} & \textbf{1{,}676.0} & \textbf{74.0} & \textbf{58.3} & \textbf{17.7} & \textbf{9.4} & \textbf{0.605} & \textbf{11.14} & \textbf{0.830} \\
TabPFN-3 (default) & 1{,}625.9 & 61.1 & 44.6 & 20.8 & 13.3 & 0.534 & 11.71 & 0.798 \\
Xiaomi-TabLDM (default) & 1{,}583.4 & 59.7 & 55.8 & 23.7 & 12.9 & 0.507 & 12.20 & 0.768 \\
TabPFN-2.6 (default) & 1{,}577.2 & 53.9 & 36.9 & 24.2 & 18.1 & 0.458 & 12.98 & 0.763 \\
RealTabPFN-2.5 (tuned + ensembled) & 1{,}561.3 & 55.3 & 44.0 & 25.4 & 16.3 & 0.425 & 12.70 & 0.751 \\
TabICLv2 (default) & 1{,}561.0 & 55.8 & 52.0 & 25.4 & 14.5 & 0.488 & 12.47 & 0.751 \\
RealTabPFN-2.5 (tuned) & 1{,}527.3 & 50.4 & 40.9 & 28.1 & 17.9 & 0.390 & 13.37 & 0.724 \\
TabDPT-1.3 (default) & 1{,}517.2 & 57.4 & 41.1 & 28.9 & 14.8 & 0.404 & 14.21 & 0.715 \\
RealTabPFN-2.5 (default) & 1{,}498.9 & 49.3 & 39.9 & 30.5 & 21.7 & 0.366 & 13.92 & 0.699 \\
FTFM (default) & 1{,}484.3 & 93.0 & 97.9 & 31.7 & 11.0 & 0.463 & 13.35 & 0.686 \\
RealMLP (tuned + ensembled) & 1{,}475.6 & 42.9 & 41.1 & 32.5 & 23.4 & 0.293 & 15.07 & 0.678 \\
TabDPT-Turbo (default) & 1{,}433.3 & 49.9 & 42.7 & 36.5 & 20.9 & 0.292 & 16.09 & 0.638 \\
TabDPT (tuned + ensembled) & 1{,}431.4 & 50.3 & 43.8 & 36.6 & 22.0 & 0.276 & 15.93 & 0.636 \\
TabM (tuned + ensembled) & 1{,}422.1 & 42.7 & 35.5 & 37.5 & 27.4 & 0.220 & 16.19 & 0.627 \\
LightGBM (tuned + ensembled) & 1{,}410.5 & 26.9 & 26.3 & 38.7 & 32.1 & 0.165 & 17.11 & 0.616 \\
RealMLP (tuned) & 1{,}406.0 & 35.1 & 45.1 & 39.1 & 26.3 & 0.207 & 16.32 & 0.611 \\
CatBoost (tuned + ensembled) & 1{,}400.0 & 36.6 & 35.5 & 39.7 & 26.5 & 0.171 & 16.73 & 0.605 \\
CatBoost (tuned) & 1{,}388.8 & 30.5 & 31.4 & 40.8 & 26.8 & 0.157 & 16.94 & 0.593 \\
iLTM (tuned + ensembled) & 1{,}383.5 & 36.1 & 38.6 & 41.4 & 32.1 & 0.158 & 17.37 & 0.588 \\
TabDPT (tuned) & 1{,}382.7 & 50.3 & 50.7 & 41.5 & 26.3 & 0.242 & 17.01 & 0.587 \\
TabM (tuned) & 1{,}372.2 & 43.9 & 36.1 & 42.5 & 30.1 & 0.171 & 17.03 & 0.576 \\
LightGBM (tuned) & 1{,}367.6 & 25.8 & 25.5 & 43.0 & 36.2 & 0.114 & 17.74 & 0.571 \\
ModernNCA (tuned + ensembled) & 1{,}367.5 & 61.1 & 48.9 & 43.0 & 27.7 & 0.201 & 17.66 & 0.571 \\
ChimeraBoost (tuned + ensembled) & 1{,}364.7 & 41.4 & 48.9 & 43.3 & 21.5 & 0.163 & 18.01 & 0.568 \\
CatBoost (default) & 1{,}360.9 & 38.1 & 37.7 & 43.7 & 26.8 & 0.142 & 17.56 & 0.564 \\
XGBoost (tuned + ensembled) & 1{,}358.9 & 28.7 & 29.0 & 43.9 & 35.5 & 0.120 & 17.82 & 0.562 \\
LimiX (default) & 1{,}344.5 & 73.6 & 56.2 & 45.4 & 24.8 & 0.231 & 17.58 & 0.547 \\
ModernNCA (tuned) & 1{,}340.0 & 34.1 & 36.8 & 45.9 & 28.1 & 0.132 & 18.06 & 0.542 \\
ChimeraBoost (tuned) & 1{,}338.4 & 43.0 & 48.9 & 46.1 & 25.8 & 0.137 & 18.45 & 0.540 \\
XGBoost (tuned) & 1{,}336.0 & 27.2 & 30.1 & 46.3 & 38.4 & 0.095 & 18.12 & 0.538 \\
xRFM (tuned + ensembled) & 1{,}335.9 & 43.4 & 44.2 & 46.3 & 30.5 & 0.152 & 18.32 & 0.538 \\
TabSwift (default) & 1{,}332.4 & 59.0 & 46.6 & 46.7 & 34.4 & 0.179 & 18.16 & 0.534 \\
CTBoost (tuned + ensembled) & 1{,}326.1 & 40.3 & 45.5 & 47.3 & 32.3 & 0.127 & 18.75 & 0.527 \\
TabPFN-v2 (tuned + ensembled) & 1{,}325.8 & 61.4 & 65.7 & 47.4 & 26.0 & 0.234 & 18.67 & 0.527 \\
LightPFN (default) & 1{,}322.6 & 65.1 & 66.8 & 47.7 & 28.7 & 0.234 & 18.61 & 0.523 \\
Mitra (default) & 1{,}315.6 & 59.7 & 61.9 & 48.5 & 28.6 & 0.226 & 19.18 & 0.516 \\
TabDPT (default) & 1{,}311.9 & 60.3 & 54.5 & 48.9 & 30.0 & 0.190 & 19.19 & 0.512 \\
TabICL (default) & 1{,}303.3 & 50.1 & 55.5 & 49.8 & 34.5 & 0.172 & 19.08 & 0.502 \\
ChimeraBoost (default) & 1{,}302.7 & 43.1 & 55.9 & 49.8 & 30.8 & 0.118 & 19.26 & 0.502 \\
CTBoost (tuned) & 1{,}296.6 & 38.7 & 47.3 & 50.5 & 37.9 & 0.098 & 19.28 & 0.495 \\
xRFM (tuned) & 1{,}292.1 & 43.6 & 42.9 & 51.0 & 34.8 & 0.109 & 19.43 & 0.490 \\
iLTM (tuned) & 1{,}284.3 & 32.8 & 34.7 & 51.8 & 39.2 & 0.070 & 19.14 & 0.481 \\
TabM (default) & 1{,}283.9 & 42.1 & 40.8 & 51.9 & 38.8 & 0.100 & 19.02 & 0.481 \\
TabPFN-v2 (tuned) & 1{,}272.0 & 54.8 & 56.1 & 53.1 & 29.8 & 0.174 & 20.11 & 0.468 \\
NN-Torch (tuned + ensembled) & 1{,}270.7 & 45.4 & 46.8 & 53.3 & 43.4 & 0.079 & 19.19 & 0.467 \\
Beta (default) & 1{,}269.8 & 58.7 & 60.3 & 53.4 & 34.8 & 0.124 & 20.47 & 0.466 \\
SAP-RPT-OSS (default) & 1{,}268.7 & 60.6 & 58.1 & 53.5 & 23.8 & 0.140 & 20.30 & 0.464 \\
EBM (tuned + ensembled) & 1{,}254.2 & 35.6 & 39.2 & 55.0 & 42.0 & 0.062 & 20.62 & 0.449 \\
TabPFN-v2 (default) & 1{,}245.9 & 63.1 & 62.1 & 55.9 & 31.3 & 0.153 & 20.91 & 0.440 \\
ModernNCA (default) & 1{,}241.7 & 37.4 & 37.2 & 56.4 & 37.5 & 0.054 & 21.11 & 0.435 \\
EBM (tuned) & 1{,}220.4 & 46.2 & 42.2 & 58.6 & 45.0 & 0.041 & 21.24 & 0.412 \\
RealMLP (default) & 1{,}220.0 & 32.6 & 41.3 & 58.7 & 44.6 & 0.038 & 20.52 & 0.411 \\
PrismBoost (tuned + ensembled) & 1{,}215.5 & 48.3 & 46.5 & 59.1 & 48.6 & 0.053 & 22.58 & 0.407 \\
XGBoost (default) & 1{,}209.0 & 31.8 & 36.6 & 59.8 & 51.3 & 0.031 & 20.77 & 0.400 \\
ExtraTrees (tuned + ensembled) & 1{,}204.3 & 44.8 & 46.4 & 60.3 & 42.2 & 0.066 & 21.68 & 0.395 \\
NN-Torch (tuned) & 1{,}201.9 & 43.1 & 45.4 & 60.6 & 48.2 & 0.041 & 20.83 & 0.392 \\
FastaiMLP (tuned + ensembled) & 1{,}193.9 & 54.6 & 59.9 & 61.4 & 44.4 & 0.067 & 21.58 & 0.384 \\
EBM (default) & 1{,}191.5 & 47.0 & 51.0 & 61.7 & 43.1 & 0.041 & 22.15 & 0.381 \\
APLR (tuned + ensembled) & 1{,}185.0 & 41.3 & 61.1 & 62.3 & 49.5 & 0.044 & 22.25 & 0.374 \\
LightGBM (default) & 1{,}183.8 & 34.7 & 33.3 & 62.5 & 55.3 & 0.023 & 21.27 & 0.373 \\
PrismBoost (tuned) & 1{,}182.4 & 42.2 & 42.6 & 62.6 & 50.7 & 0.037 & 23.13 & 0.371 \\
ExtraTrees (tuned) & 1{,}180.5 & 42.8 & 55.8 & 62.8 & 37.6 & 0.054 & 22.42 & 0.370 \\
TabSTAR-v1.1 (default) & 1{,}179.0 & 65.2 & 84.7 & 62.9 & 33.5 & 0.089 & 23.52 & 0.368 \\
RandomForest (tuned + ensembled) & 1{,}173.3 & 50.9 & 46.2 & 63.5 & 41.7 & 0.050 & 22.66 & 0.362 \\
CTBoost (default) & 1{,}173.1 & 62.0 & 67.2 & 63.6 & 42.6 & 0.070 & 23.15 & 0.362 \\
Nori-30M (default) & 1{,}161.5 & 73.0 & 84.0 & 64.7 & 28.9 & 0.139 & 24.91 & 0.350 \\
Nori (default) & 1{,}151.9 & 70.5 & 79.4 & 65.7 & 39.4 & 0.122 & 25.15 & 0.340 \\
APLR (tuned) & 1{,}143.6 & 46.0 & 58.4 & 66.5 & 53.8 & 0.024 & 23.34 & 0.331 \\
FastaiMLP (tuned) & 1{,}138.4 & 55.2 & 61.4 & 67.0 & 48.4 & 0.025 & 22.94 & 0.326 \\
RandomForest (tuned) & 1{,}137.9 & 47.9 & 48.3 & 67.1 & 39.2 & 0.031 & 23.35 & 0.326 \\
PrismBoost (default) & 1{,}126.6 & 47.3 & 52.6 & 68.2 & 57.1 & 0.031 & 24.77 & 0.315 \\
PerpetualBooster (tuned + ensembled) & 1{,}093.2 & 43.1 & 46.9 & 71.3 & 60.3 & 0.017 & 28.01 & 0.282 \\
Orion-MSP (default) & 1{,}085.4 & 53.4 & 54.6 & 72.0 & 56.4 & 0.034 & 25.44 & 0.275 \\
iLTM (default) & 1{,}084.1 & 46.9 & 59.6 & 72.2 & 50.5 & 0.021 & 24.83 & 0.274 \\
APLR (default) & 1{,}076.4 & 51.8 & 83.2 & 72.8 & 61.2 & 0.014 & 26.21 & 0.267 \\
NN-Torch (default) & 1{,}072.6 & 39.4 & 50.4 & 73.2 & 59.4 & 0.004 & 24.62 & 0.263 \\
PerpetualBooster (tuned) & 1{,}058.2 & 43.3 & 45.8 & 74.4 & 57.2 & 0.007 & 29.29 & 0.251 \\
xRFM (default) & 1{,}044.2 & 55.8 & 68.8 & 75.6 & 66.1 & 0.020 & 28.15 & 0.238 \\
ExtraTrees (default) & 1{,}010.7 & 45.7 & 62.0 & 78.3 & 69.9 & 0.008 & 27.86 & 0.211 \\
TabFlex (default) & 1{,}010.1 & 66.8 & 72.9 & 78.4 & 57.2 & 0.026 & 29.28 & 0.210 \\
FastaiMLP (default) & 1{,}005.4 & 60.8 & 64.0 & 78.7 & 59.7 & 0.005 & 27.36 & 0.207 \\
RandomForest (default) & 1{,}000.0 & 41.8 & 45.4 & 79.2 & 72.8 & 0.000 & 27.95 & 0.202 \\
KNN (tuned + ensembled) & 992.7 & 63.6 & 82.0 & 79.7 & 56.2 & 0.010 & 29.43 & 0.197 \\
Linear (tuned + ensembled) & 987.8 & 70.6 & 103.8 & 80.1 & 52.9 & 0.029 & 33.78 & 0.193 \\
Linear (tuned) & 948.8 & 68.2 & 116.5 & 82.8 & 54.9 & 0.020 & 34.58 & 0.165 \\
PerpetualBooster (default) & 935.4 & 41.0 & 66.5 & 83.7 & 63.1 & 0.000 & 33.14 & 0.157 \\
KNN (tuned) & 887.8 & 59.9 & 100.1 & 86.5 & 65.7 & 0.003 & 34.21 & 0.128 \\
Linear (default) & 861.1 & 67.4 & 121.2 & 87.9 & 67.5 & 0.009 & 38.20 & 0.113 \\
KNN (default) & 631.2 & 88.0 & 103.4 & 95.5 & 87.5 & 0.000 & 47.05 & 0.035 \\
\end{longtable}
}

{\scriptsize
\setlength{\tabcolsep}{3.2pt}
\begin{longtable}{l r r r r r r r r}
\caption{Full TabArena regression leaderboard: all 13 regression datasets, 99 evaluated methods, sorted by Elo.}\label{tab:app-regression}\\
\toprule
Method & Elo & Elo$+$ & Elo$-$ & Avg.\ rank & Harm.\ rank & Score & Improv.\ (\%) & Winrate \\
\midrule
\endfirsthead
\multicolumn{9}{l}{\footnotesize\emph{Table \ref{tab:app-regression} (continued)}}\\
\toprule
Method & Elo & Elo$+$ & Elo$-$ & Avg.\ rank & Harm.\ rank & Score & Improv.\ (\%) & Winrate \\
\midrule
\endhead
\midrule
\multicolumn{9}{r}{\footnotesize continued on next page}\\
\midrule
\endfoot
\bottomrule
\endlastfoot
Kumo-Tabular (default) & 2{,}394.7 & 264.2 & 183.1 & 2.9 & 1.7 & 0.969 & 0.70 & 0.980 \\
LimiX-2 (default) & 2{,}269.0 & 329.2 & 173.5 & 4.6 & 1.9 & 0.931 & 0.95 & 0.963 \\
Kumo-Tabular-Medium (default) & 2{,}249.8 & 208.3 & 136.4 & 4.9 & 3.6 & 0.896 & 1.72 & 0.960 \\
TabPFN-3.5 (default) & 2{,}144.3 & 220.0 & 117.0 & 7.0 & 4.2 & 0.826 & 2.75 & 0.939 \\
Kumo-Tabular-Small (default) & 2{,}117.9 & 150.7 & 84.3 & 7.6 & 5.5 & 0.815 & 2.71 & 0.933 \\
Causilo (default) & 2{,}038.6 & 176.3 & 109.7 & 9.7 & 6.1 & 0.768 & 3.27 & 0.911 \\
\textbf{TabAbax} & \textbf{2{,}023.7} & \textbf{179.7} & \textbf{95.8} & \textbf{10.1} & \textbf{6.3} & \textbf{0.749} & \textbf{3.30} & \textbf{0.907} \\
TabFM (default) & 2{,}002.9 & 140.9 & 78.5 & 10.8 & 7.2 & 0.733 & 4.09 & 0.900 \\
Mitra-v2 (default) & 1{,}987.5 & 170.2 & 106.9 & 11.2 & 7.2 & 0.726 & 3.33 & 0.896 \\
TabPFN-3.5-Fast (default) & 1{,}949.6 & 146.5 & 82.7 & 12.5 & 7.6 & 0.703 & 4.59 & 0.883 \\
EXAONE-Tabular (default) & 1{,}895.9 & 118.3 & 79.9 & 14.5 & 10.9 & 0.650 & 5.15 & 0.862 \\
TabPFN-3 (default) & 1{,}807.2 & 187.6 & 113.8 & 18.3 & 13.2 & 0.585 & 4.88 & 0.824 \\
Nori-30M (default) & 1{,}778.5 & 130.4 & 72.7 & 19.6 & 10.5 & 0.544 & 5.93 & 0.810 \\
Xiaomi-TabLDM (default) & 1{,}777.1 & 233.1 & 174.4 & 19.7 & 12.3 & 0.602 & 5.26 & 0.809 \\
TabPFN-2.6 (default) & 1{,}751.1 & 93.0 & 39.1 & 21.0 & 17.9 & 0.518 & 6.40 & 0.796 \\
RealTabPFN-2.5 (tuned + ensembled) & 1{,}750.0 & 122.7 & 88.0 & 21.1 & 16.2 & 0.491 & 5.82 & 0.795 \\
TabDPT (tuned + ensembled) & 1{,}741.6 & 132.2 & 71.4 & 21.5 & 14.6 & 0.490 & 6.61 & 0.791 \\
Nori (default) & 1{,}705.9 & 133.8 & 75.5 & 23.4 & 16.9 & 0.478 & 6.89 & 0.771 \\
TabDPT (tuned) & 1{,}690.0 & 101.3 & 56.2 & 24.3 & 19.0 & 0.454 & 7.03 & 0.762 \\
TabICLv2 (default) & 1{,}688.1 & 218.7 & 130.4 & 24.4 & 15.5 & 0.525 & 6.35 & 0.761 \\
TabDPT-1.3 (default) & 1{,}672.3 & 193.7 & 115.4 & 25.3 & 14.4 & 0.458 & 6.87 & 0.752 \\
RealMLP (tuned + ensembled) & 1{,}670.3 & 80.1 & 48.5 & 25.5 & 20.8 & 0.400 & 7.40 & 0.750 \\
RealTabPFN-2.5 (tuned) & 1{,}643.9 & 115.5 & 94.0 & 27.0 & 19.5 & 0.403 & 6.55 & 0.734 \\
TabDPT-Turbo (default) & 1{,}624.9 & 170.0 & 77.3 & 28.2 & 16.0 & 0.390 & 7.53 & 0.722 \\
TabDPT (default) & 1{,}607.5 & 133.0 & 61.3 & 29.3 & 20.1 & 0.377 & 7.91 & 0.711 \\
RealMLP (tuned) & 1{,}576.6 & 86.5 & 61.5 & 31.4 & 26.4 & 0.295 & 8.33 & 0.690 \\
ModernNCA (tuned + ensembled) & 1{,}558.9 & 114.8 & 106.0 & 32.6 & 21.3 & 0.294 & 9.49 & 0.678 \\
RealTabPFN-2.5 (default) & 1{,}553.0 & 128.3 & 123.6 & 33.0 & 26.0 & 0.333 & 7.93 & 0.674 \\
CatBoost (tuned + ensembled) & 1{,}520.5 & 78.0 & 53.4 & 35.3 & 30.9 & 0.197 & 10.18 & 0.650 \\
LightGBM (tuned + ensembled) & 1{,}514.5 & 66.6 & 63.8 & 35.7 & 32.2 & 0.183 & 10.59 & 0.646 \\
CatBoost (tuned) & 1{,}496.9 & 85.5 & 60.8 & 37.0 & 32.4 & 0.179 & 10.32 & 0.632 \\
xRFM (tuned + ensembled) & 1{,}495.1 & 90.4 & 94.2 & 37.2 & 29.4 & 0.221 & 9.71 & 0.631 \\
TabM (tuned + ensembled) & 1{,}475.9 & 118.4 & 84.1 & 38.6 & 33.1 & 0.214 & 9.20 & 0.616 \\
iLTM (tuned + ensembled) & 1{,}467.7 & 66.5 & 49.4 & 39.2 & 33.8 & 0.169 & 11.49 & 0.610 \\
LightGBM (tuned) & 1{,}451.0 & 72.5 & 79.0 & 40.5 & 36.2 & 0.134 & 11.15 & 0.597 \\
ChimeraBoost (tuned + ensembled) & 1{,}443.8 & 114.2 & 99.5 & 41.1 & 13.2 & 0.172 & 11.44 & 0.591 \\
XGBoost (tuned + ensembled) & 1{,}443.8 & 51.5 & 44.8 & 41.1 & 38.4 & 0.117 & 11.11 & 0.591 \\
XGBoost (tuned) & 1{,}421.6 & 60.0 & 56.4 & 42.8 & 39.6 & 0.102 & 11.25 & 0.574 \\
TabSwift (default) & 1{,}418.0 & 149.2 & 119.9 & 43.1 & 33.6 & 0.219 & 9.58 & 0.571 \\
xRFM (tuned) & 1{,}412.9 & 74.4 & 81.6 & 43.5 & 38.7 & 0.158 & 10.36 & 0.566 \\
ChimeraBoost (tuned) & 1{,}409.7 & 102.3 & 114.1 & 43.7 & 17.8 & 0.148 & 11.76 & 0.564 \\
CatBoost (default) & 1{,}408.6 & 86.0 & 92.6 & 43.8 & 37.5 & 0.116 & 11.71 & 0.563 \\
TabM (tuned) & 1{,}401.0 & 108.0 & 95.7 & 44.4 & 37.9 & 0.147 & 10.08 & 0.557 \\
ModernNCA (tuned) & 1{,}397.9 & 99.8 & 89.5 & 44.7 & 35.4 & 0.154 & 11.38 & 0.554 \\
iLTM (tuned) & 1{,}361.1 & 68.8 & 56.7 & 47.7 & 42.6 & 0.085 & 11.46 & 0.524 \\
ChimeraBoost (default) & 1{,}344.9 & 106.9 & 120.6 & 49.0 & 25.4 & 0.109 & 12.49 & 0.510 \\
TabPFN-v2 (tuned + ensembled) & 1{,}326.5 & 175.5 & 123.6 & 50.5 & 28.3 & 0.201 & 10.95 & 0.495 \\
PrismBoost (tuned + ensembled) & 1{,}315.3 & 68.3 & 64.4 & 51.5 & 48.7 & 0.054 & 13.06 & 0.485 \\
TabM (default) & 1{,}299.8 & 123.4 & 103.9 & 52.8 & 46.4 & 0.081 & 11.64 & 0.472 \\
ModernNCA (default) & 1{,}299.0 & 80.2 & 70.0 & 52.8 & 46.5 & 0.074 & 12.84 & 0.471 \\
Mitra (default) & 1{,}276.5 & 141.8 & 128.6 & 54.7 & 38.6 & 0.142 & 12.46 & 0.452 \\
SAP-RPT-OSS (default) & 1{,}272.3 & 165.0 & 161.6 & 55.1 & 30.5 & 0.133 & 12.94 & 0.448 \\
LimiX (default) & 1{,}258.8 & 187.9 & 155.7 & 56.2 & 38.3 & 0.120 & 12.51 & 0.437 \\
PrismBoost (tuned) & 1{,}253.2 & 60.4 & 55.7 & 56.7 & 53.9 & 0.025 & 13.72 & 0.432 \\
NN-Torch (tuned + ensembled) & 1{,}249.2 & 114.2 & 124.7 & 57.0 & 45.9 & 0.068 & 13.09 & 0.429 \\
CTBoost (tuned + ensembled) & 1{,}247.3 & 96.0 & 137.9 & 57.2 & 50.4 & 0.049 & 14.64 & 0.427 \\
ExtraTrees (tuned + ensembled) & 1{,}246.7 & 104.2 & 106.2 & 57.2 & 47.0 & 0.083 & 14.94 & 0.426 \\
TabPFN-v2 (tuned) & 1{,}239.7 & 157.2 & 147.6 & 57.8 & 38.8 & 0.127 & 11.96 & 0.420 \\
RealMLP (default) & 1{,}238.5 & 110.1 & 93.6 & 57.9 & 51.2 & 0.063 & 12.70 & 0.420 \\
LightGBM (default) & 1{,}232.7 & 37.2 & 52.4 & 58.4 & 56.5 & 0.006 & 13.46 & 0.415 \\
ExtraTrees (tuned) & 1{,}225.6 & 104.2 & 104.9 & 59.0 & 44.1 & 0.072 & 15.16 & 0.409 \\
PrismBoost (default) & 1{,}218.3 & 75.0 & 71.2 & 59.6 & 54.8 & 0.027 & 14.13 & 0.402 \\
XGBoost (default) & 1{,}211.8 & 79.7 & 88.5 & 60.1 & 55.3 & 0.025 & 14.09 & 0.397 \\
CTBoost (tuned) & 1{,}206.2 & 100.8 & 136.8 & 60.6 & 54.2 & 0.035 & 15.01 & 0.392 \\
PerpetualBooster (tuned + ensembled) & 1{,}195.1 & 67.3 & 100.1 & 61.5 & 57.9 & 0.012 & 15.14 & 0.383 \\
TabPFN-v2 (default) & 1{,}194.8 & 139.8 & 162.0 & 61.5 & 49.0 & 0.104 & 13.30 & 0.383 \\
NN-Torch (tuned) & 1{,}192.8 & 102.0 & 118.2 & 61.7 & 53.1 & 0.047 & 13.85 & 0.381 \\
RandomForest (tuned + ensembled) & 1{,}182.8 & 68.4 & 93.7 & 62.5 & 59.3 & 0.019 & 15.91 & 0.372 \\
xRFM (default) & 1{,}161.2 & 124.0 & 122.9 & 64.2 & 55.9 & 0.034 & 15.74 & 0.355 \\
EBM (tuned + ensembled) & 1{,}149.5 & 144.3 & 189.7 & 65.2 & 54.1 & 0.051 & 16.41 & 0.345 \\
TabSTAR-v1.1 (default) & 1{,}143.8 & 121.7 & 145.8 & 65.6 & 58.1 & 0.036 & 16.09 & 0.340 \\
PerpetualBooster (tuned) & 1{,}140.2 & 56.6 & 102.0 & 65.9 & 62.9 & 0.000 & 16.16 & 0.337 \\
RandomForest (tuned) & 1{,}130.4 & 69.1 & 106.2 & 66.7 & 63.3 & 0.007 & 16.37 & 0.330 \\
EBM (tuned) & 1{,}104.5 & 144.2 & 184.9 & 68.7 & 57.2 & 0.034 & 16.93 & 0.309 \\
ExtraTrees (default) & 1{,}065.6 & 91.2 & 118.9 & 71.6 & 63.4 & 0.012 & 17.50 & 0.279 \\
EBM (default) & 1{,}043.2 & 128.7 & 186.5 & 73.3 & 64.8 & 0.017 & 17.81 & 0.262 \\
NN-Torch (default) & 1{,}043.0 & 90.6 & 124.9 & 73.3 & 68.7 & 0.002 & 17.11 & 0.262 \\
APLR (tuned + ensembled) & 1{,}036.4 & 111.3 & 135.4 & 73.8 & 69.2 & 0.001 & 17.87 & 0.258 \\
FastaiMLP (tuned + ensembled) & 1{,}025.6 & 111.7 & 132.4 & 74.5 & 69.6 & 0.015 & 17.30 & 0.250 \\
RandomForest (default) & 1{,}000.0 & 45.5 & 89.2 & 76.3 & 74.4 & 0.000 & 17.84 & 0.232 \\
FTFM (default) & 1{,}000.0 & 45.5 & 89.2 & 76.3 & 74.4 & 0.000 & 17.84 & 0.232 \\
LightPFN (default) & 1{,}000.0 & 45.5 & 89.2 & 76.3 & 74.4 & 0.000 & 17.84 & 0.232 \\
Beta (default) & 1{,}000.0 & 45.5 & 89.2 & 76.3 & 74.4 & 0.000 & 17.84 & 0.232 \\
TabFlex (default) & 1{,}000.0 & 45.5 & 89.2 & 76.3 & 74.4 & 0.000 & 17.84 & 0.232 \\
Orion-MSP (default) & 1{,}000.0 & 45.5 & 89.2 & 76.3 & 74.4 & 0.000 & 17.84 & 0.232 \\
TabICL (default) & 1{,}000.0 & 45.5 & 89.2 & 76.3 & 74.4 & 0.000 & 17.84 & 0.232 \\
APLR (tuned) & 993.2 & 110.9 & 143.8 & 76.7 & 71.2 & 0.002 & 18.63 & 0.228 \\
FastaiMLP (tuned) & 981.1 & 115.3 & 135.5 & 77.5 & 73.4 & 0.009 & 17.77 & 0.219 \\
CTBoost (default) & 979.2 & 114.6 & 185.9 & 77.6 & 72.0 & 0.014 & 20.26 & 0.218 \\
PerpetualBooster (default) & 929.9 & 94.4 & 138.2 & 80.6 & 78.5 & 0.000 & 19.53 & 0.187 \\
KNN (tuned + ensembled) & 873.3 & 160.8 & 194.6 & 83.7 & 72.0 & 0.000 & 22.69 & 0.156 \\
FastaiMLP (default) & 860.9 & 124.4 & 169.2 & 84.3 & 81.8 & 0.000 & 21.72 & 0.150 \\
iLTM (default) & 852.6 & 79.4 & 122.3 & 84.7 & 83.6 & 0.000 & 20.75 & 0.146 \\
APLR (default) & 839.8 & 130.9 & 214.6 & 85.3 & 81.6 & 0.000 & 23.00 & 0.139 \\
KNN (tuned) & 767.1 & 171.2 & 211.1 & 88.4 & 81.0 & 0.000 & 24.97 & 0.108 \\
KNN (default) & 655.2 & 201.1 & 339.4 & 91.9 & 72.7 & 0.000 & 31.81 & 0.072 \\
Linear (tuned + ensembled) & 434.8 & 128.7 & 371.6 & 96.0 & 95.8 & 0.000 & 38.65 & 0.031 \\
Linear (tuned) & 397.4 & 153.9 & 480.5 & 96.4 & 96.2 & 0.000 & 38.81 & 0.026 \\
Linear (default) & 183.5 & 173.4 & 460.3 & 98.2 & 98.1 & 0.000 & 41.14 & 0.008 \\
\end{longtable}
}

{\scriptsize
\setlength{\tabcolsep}{3.2pt}
\begin{longtable}{l r r r r r r r r}
\caption{Full TabArena classification leaderboard: all 38 classification datasets, 99 evaluated methods, sorted by Elo.}\label{tab:app-classification}\\
\toprule
Method & Elo & Elo$+$ & Elo$-$ & Avg.\ rank & Harm.\ rank & Score & Improv.\ (\%) & Winrate \\
\midrule
\endfirsthead
\multicolumn{9}{l}{\footnotesize\emph{Table \ref{tab:app-classification} (continued)}}\\
\toprule
Method & Elo & Elo$+$ & Elo$-$ & Avg.\ rank & Harm.\ rank & Score & Improv.\ (\%) & Winrate \\
\midrule
\endhead
\midrule
\multicolumn{9}{r}{\footnotesize continued on next page}\\
\midrule
\endfoot
\bottomrule
\endlastfoot
LimiX-2 (default) & 1{,}929.0 & 135.9 & 77.4 & 7.1 & 2.3 & 0.893 & 4.13 & 0.937 \\
Kumo-Tabular (default) & 1{,}920.4 & 158.7 & 110.1 & 7.4 & 2.7 & 0.856 & 5.65 & 0.935 \\
Kumo-Tabular-Medium (default) & 1{,}873.0 & 133.6 & 94.2 & 8.9 & 4.5 & 0.803 & 7.31 & 0.919 \\
TabPFN-3.5 (default) & 1{,}845.4 & 106.7 & 69.6 & 10.0 & 4.2 & 0.770 & 9.20 & 0.909 \\
TabPFN-3.5-Fast (default) & 1{,}779.8 & 95.1 & 69.4 & 12.8 & 6.9 & 0.701 & 10.45 & 0.880 \\
TabFM (default) & 1{,}778.4 & 125.3 & 117.3 & 12.8 & 3.1 & 0.740 & 7.42 & 0.879 \\
Kumo-Tabular-Small (default) & 1{,}775.8 & 88.4 & 55.7 & 12.9 & 8.6 & 0.689 & 11.01 & 0.878 \\
Causilo (default) & 1{,}769.9 & 81.6 & 64.1 & 13.2 & 7.4 & 0.688 & 11.05 & 0.875 \\
Mitra-v2 (default) & 1{,}756.6 & 122.3 & 85.5 & 13.9 & 5.7 & 0.692 & 10.09 & 0.869 \\
EXAONE-Tabular (default) & 1{,}754.3 & 85.4 & 52.8 & 14.0 & 7.0 & 0.683 & 11.19 & 0.867 \\
FTFM (default) & 1{,}707.1 & 79.1 & 89.2 & 16.5 & 8.5 & 0.621 & 11.82 & 0.842 \\
\textbf{TabAbax} & \textbf{1{,}646.7} & \textbf{81.7} & \textbf{72.4} & \textbf{20.2} & \textbf{11.2} & \textbf{0.555} & \textbf{13.83} & \textbf{0.804} \\
TabPFN-3 (default) & 1{,}626.0 & 72.7 & 63.9 & 21.6 & 13.4 & 0.516 & 14.04 & 0.789 \\
Xiaomi-TabLDM (default) & 1{,}578.7 & 70.4 & 54.2 & 25.1 & 13.1 & 0.475 & 14.57 & 0.754 \\
TabPFN-2.6 (default) & 1{,}576.6 & 53.5 & 48.2 & 25.3 & 18.2 & 0.437 & 15.24 & 0.752 \\
TabICLv2 (default) & 1{,}570.6 & 75.6 & 64.0 & 25.7 & 14.1 & 0.475 & 14.56 & 0.748 \\
RealTabPFN-2.5 (tuned + ensembled) & 1{,}556.5 & 62.5 & 60.8 & 26.9 & 16.3 & 0.403 & 15.05 & 0.736 \\
RealTabPFN-2.5 (tuned) & 1{,}537.3 & 61.4 & 53.6 & 28.4 & 17.4 & 0.386 & 15.71 & 0.720 \\
RealTabPFN-2.5 (default) & 1{,}523.5 & 53.8 & 49.8 & 29.6 & 20.5 & 0.378 & 15.97 & 0.708 \\
TabDPT-1.3 (default) & 1{,}517.3 & 73.8 & 55.1 & 30.1 & 15.0 & 0.386 & 16.72 & 0.703 \\
RealMLP (tuned + ensembled) & 1{,}464.3 & 48.6 & 41.8 & 34.9 & 24.5 & 0.256 & 17.69 & 0.654 \\
TabM (tuned + ensembled) & 1{,}440.9 & 51.8 & 39.2 & 37.2 & 25.8 & 0.222 & 18.58 & 0.631 \\
LightPFN (default) & 1{,}432.9 & 68.2 & 79.1 & 38.0 & 23.7 & 0.314 & 18.87 & 0.623 \\
TabDPT-Turbo (default) & 1{,}419.7 & 70.3 & 54.4 & 39.3 & 23.4 & 0.259 & 19.02 & 0.609 \\
LightGBM (tuned + ensembled) & 1{,}415.6 & 44.5 & 30.0 & 39.7 & 32.0 & 0.159 & 19.33 & 0.605 \\
TabICL (default) & 1{,}405.3 & 49.0 & 51.4 & 40.7 & 29.2 & 0.231 & 19.50 & 0.595 \\
CatBoost (tuned + ensembled) & 1{,}400.2 & 48.8 & 43.2 & 41.2 & 25.3 & 0.163 & 18.96 & 0.589 \\
LimiX (default) & 1{,}395.5 & 82.0 & 70.8 & 41.7 & 22.1 & 0.269 & 19.31 & 0.585 \\
RealMLP (tuned) & 1{,}395.0 & 55.4 & 45.7 & 41.8 & 26.3 & 0.176 & 19.05 & 0.584 \\
TabDPT (tuned + ensembled) & 1{,}394.5 & 64.6 & 45.6 & 41.8 & 26.5 & 0.203 & 19.12 & 0.583 \\
TabM (tuned) & 1{,}393.8 & 46.8 & 40.1 & 41.9 & 28.1 & 0.180 & 19.40 & 0.583 \\
iLTM (tuned + ensembled) & 1{,}391.6 & 42.2 & 39.3 & 42.1 & 31.6 & 0.155 & 19.38 & 0.580 \\
CatBoost (tuned) & 1{,}391.3 & 46.3 & 43.0 & 42.1 & 25.2 & 0.150 & 19.21 & 0.580 \\
CatBoost (default) & 1{,}376.6 & 44.8 & 43.6 & 43.7 & 24.4 & 0.151 & 19.56 & 0.565 \\
LightGBM (tuned) & 1{,}374.7 & 35.2 & 31.0 & 43.9 & 36.2 & 0.107 & 20.00 & 0.563 \\
CTBoost (tuned + ensembled) & 1{,}373.4 & 50.2 & 55.7 & 44.0 & 28.8 & 0.154 & 20.16 & 0.561 \\
ChimeraBoost (tuned + ensembled) & 1{,}372.6 & 48.7 & 57.2 & 44.1 & 27.5 & 0.160 & 20.26 & 0.560 \\
XGBoost (tuned + ensembled) & 1{,}364.9 & 41.6 & 45.7 & 44.9 & 34.7 & 0.121 & 20.12 & 0.552 \\
Beta (default) & 1{,}358.6 & 48.2 & 49.0 & 45.5 & 29.5 & 0.167 & 21.37 & 0.546 \\
ModernNCA (tuned) & 1{,}351.4 & 50.0 & 36.3 & 46.3 & 26.2 & 0.124 & 20.34 & 0.538 \\
TabPFN-v2 (tuned + ensembled) & 1{,}351.4 & 85.9 & 71.9 & 46.3 & 25.3 & 0.245 & 21.31 & 0.538 \\
Mitra (default) & 1{,}351.2 & 77.2 & 77.1 & 46.3 & 26.3 & 0.255 & 21.48 & 0.537 \\
ModernNCA (tuned + ensembled) & 1{,}348.7 & 72.4 & 66.9 & 46.6 & 30.9 & 0.169 & 20.46 & 0.535 \\
ChimeraBoost (tuned) & 1{,}346.2 & 48.2 & 56.1 & 46.8 & 30.4 & 0.133 & 20.74 & 0.532 \\
CTBoost (tuned) & 1{,}344.4 & 46.7 & 53.1 & 47.0 & 34.4 & 0.119 & 20.74 & 0.530 \\
TabDPT (tuned) & 1{,}341.6 & 57.0 & 50.7 & 47.3 & 30.3 & 0.170 & 20.42 & 0.527 \\
XGBoost (tuned) & 1{,}340.0 & 36.0 & 40.7 & 47.5 & 37.9 & 0.092 & 20.48 & 0.525 \\
TabSwift (default) & 1{,}336.1 & 57.7 & 58.8 & 47.9 & 34.7 & 0.166 & 21.10 & 0.521 \\
xRFM (tuned + ensembled) & 1{,}321.8 & 51.2 & 48.5 & 49.4 & 30.9 & 0.128 & 21.26 & 0.506 \\
ChimeraBoost (default) & 1{,}315.4 & 52.9 & 65.5 & 50.1 & 33.2 & 0.121 & 21.58 & 0.499 \\
TabPFN-v2 (tuned) & 1{,}302.2 & 69.2 & 76.1 & 51.5 & 27.6 & 0.190 & 22.90 & 0.484 \\
TabM (default) & 1{,}302.2 & 51.8 & 44.2 & 51.5 & 36.7 & 0.106 & 21.54 & 0.484 \\
EBM (tuned + ensembled) & 1{,}301.9 & 40.9 & 37.2 & 51.6 & 39.0 & 0.066 & 22.06 & 0.484 \\
NN-Torch (tuned + ensembled) & 1{,}298.0 & 42.6 & 36.0 & 52.0 & 42.6 & 0.083 & 21.28 & 0.480 \\
SAP-RPT-OSS (default) & 1{,}289.0 & 55.6 & 70.8 & 52.9 & 22.2 & 0.143 & 22.82 & 0.470 \\
iLTM (tuned) & 1{,}286.4 & 42.4 & 38.9 & 53.2 & 38.1 & 0.065 & 21.77 & 0.467 \\
xRFM (tuned) & 1{,}283.5 & 56.8 & 65.8 & 53.5 & 33.6 & 0.093 & 22.53 & 0.464 \\
TabPFN-v2 (default) & 1{,}279.1 & 74.8 & 80.4 & 54.0 & 27.8 & 0.170 & 23.51 & 0.459 \\
EBM (tuned) & 1{,}268.1 & 43.2 & 38.6 & 55.2 & 42.0 & 0.043 & 22.71 & 0.447 \\
TabDPT (default) & 1{,}264.7 & 60.4 & 59.5 & 55.5 & 36.1 & 0.126 & 23.05 & 0.443 \\
FastaiMLP (tuned + ensembled) & 1{,}251.7 & 59.8 & 70.3 & 56.9 & 39.6 & 0.085 & 23.05 & 0.429 \\
ModernNCA (default) & 1{,}245.5 & 42.0 & 42.3 & 57.6 & 35.1 & 0.048 & 23.94 & 0.423 \\
EBM (default) & 1{,}244.5 & 44.5 & 51.9 & 57.7 & 38.7 & 0.050 & 23.64 & 0.422 \\
APLR (tuned + ensembled) & 1{,}237.4 & 45.6 & 60.3 & 58.4 & 45.1 & 0.058 & 23.75 & 0.414 \\
CTBoost (default) & 1{,}234.4 & 63.3 & 67.4 & 58.7 & 37.4 & 0.089 & 24.14 & 0.411 \\
RealMLP (default) & 1{,}232.5 & 41.4 & 41.7 & 58.9 & 42.7 & 0.030 & 23.20 & 0.409 \\
XGBoost (default) & 1{,}224.9 & 37.9 & 47.9 & 59.7 & 50.0 & 0.032 & 23.05 & 0.401 \\
NN-Torch (tuned) & 1{,}220.6 & 50.7 & 38.6 & 60.2 & 46.8 & 0.039 & 23.22 & 0.396 \\
ExtraTrees (tuned + ensembled) & 1{,}209.0 & 57.8 & 54.1 & 61.4 & 40.8 & 0.060 & 23.98 & 0.384 \\
PrismBoost (tuned + ensembled) & 1{,}205.3 & 67.0 & 60.4 & 61.8 & 48.5 & 0.052 & 25.84 & 0.380 \\
TabSTAR-v1.1 (default) & 1{,}202.9 & 68.6 & 74.1 & 62.0 & 29.2 & 0.107 & 26.06 & 0.377 \\
APLR (tuned) & 1{,}193.0 & 51.7 & 65.4 & 63.0 & 49.7 & 0.031 & 24.95 & 0.367 \\
FastaiMLP (tuned) & 1{,}188.9 & 58.0 & 69.7 & 63.5 & 43.3 & 0.030 & 24.71 & 0.363 \\
LightGBM (default) & 1{,}185.0 & 51.5 & 45.9 & 63.8 & 54.9 & 0.028 & 23.94 & 0.359 \\
RandomForest (tuned + ensembled) & 1{,}184.6 & 61.1 & 67.7 & 63.9 & 37.9 & 0.061 & 24.97 & 0.358 \\
ExtraTrees (tuned) & 1{,}182.5 & 64.7 & 64.3 & 64.1 & 35.9 & 0.048 & 24.90 & 0.356 \\
PrismBoost (tuned) & 1{,}177.2 & 65.2 & 59.4 & 64.6 & 49.7 & 0.041 & 26.34 & 0.351 \\
RandomForest (tuned) & 1{,}151.2 & 71.0 & 65.0 & 67.2 & 34.7 & 0.039 & 25.73 & 0.324 \\
iLTM (default) & 1{,}144.5 & 55.5 & 57.1 & 67.9 & 44.5 & 0.029 & 26.23 & 0.318 \\
APLR (default) & 1{,}137.0 & 60.4 & 71.7 & 68.6 & 56.4 & 0.019 & 27.31 & 0.310 \\
Orion-MSP (default) & 1{,}115.3 & 63.5 & 68.5 & 70.6 & 52.1 & 0.046 & 28.03 & 0.290 \\
PrismBoost (default) & 1{,}109.7 & 75.2 & 64.5 & 71.1 & 57.9 & 0.033 & 28.41 & 0.285 \\
NN-Torch (default) & 1{,}087.1 & 46.7 & 56.2 & 73.1 & 56.7 & 0.005 & 27.19 & 0.264 \\
Linear (tuned + ensembled) & 1{,}070.1 & 94.7 & 93.2 & 74.6 & 45.8 & 0.039 & 32.12 & 0.249 \\
PerpetualBooster (tuned + ensembled) & 1{,}069.3 & 58.6 & 61.6 & 74.7 & 61.2 & 0.019 & 32.41 & 0.248 \\
FastaiMLP (default) & 1{,}043.3 & 72.5 & 76.6 & 76.8 & 54.6 & 0.007 & 29.29 & 0.226 \\
PerpetualBooster (tuned) & 1{,}036.8 & 56.0 & 60.3 & 77.4 & 55.4 & 0.009 & 33.79 & 0.221 \\
Linear (tuned) & 1{,}027.0 & 90.4 & 102.2 & 78.1 & 47.8 & 0.027 & 33.13 & 0.213 \\
KNN (tuned + ensembled) & 1{,}024.3 & 74.2 & 94.1 & 78.3 & 52.3 & 0.014 & 31.73 & 0.211 \\
TabFlex (default) & 1{,}014.1 & 82.7 & 104.7 & 79.1 & 53.0 & 0.035 & 33.20 & 0.203 \\
xRFM (default) & 1{,}008.6 & 71.4 & 91.7 & 79.5 & 70.5 & 0.015 & 32.40 & 0.199 \\
RandomForest (default) & 1{,}000.0 & 60.7 & 62.4 & 80.2 & 72.2 & 0.000 & 31.40 & 0.192 \\
Nori-30M (default) & 1{,}000.0 & 60.7 & 62.4 & 80.2 & 72.2 & 0.000 & 31.40 & 0.192 \\
Nori (default) & 1{,}000.0 & 60.7 & 62.4 & 80.2 & 72.2 & 0.000 & 31.40 & 0.192 \\
ExtraTrees (default) & 993.4 & 75.0 & 73.1 & 80.6 & 72.4 & 0.007 & 31.40 & 0.187 \\
Linear (default) & 937.1 & 83.7 & 111.5 & 84.4 & 61.0 & 0.012 & 37.19 & 0.149 \\
PerpetualBooster (default) & 931.5 & 60.0 & 81.9 & 84.7 & 59.1 & 0.000 & 37.79 & 0.146 \\
KNN (tuned) & 912.1 & 89.6 & 114.1 & 85.8 & 61.7 & 0.005 & 37.37 & 0.134 \\
KNN (default) & 563.3 & 105.5 & 116.9 & 96.8 & 94.1 & 0.000 & 52.27 & 0.023 \\
\end{longtable}
}

\end{document}